%% file: main.tex
\documentclass{article}

\AddToHook{begindocument/before}{%
  \iclrfinalcopy
  \author{%
    \parbox[t]{\dimexpr\textwidth-2\tabcolsep\relax}{%
      \centering\bfseries
      Asim~Kadav, Christian~Flores, Chirag~Arora, Varun~Kotte, Hongbo~Zheng, Lan~Yan, Priya~Shanmugasundaram, Tracy~Holloway~King\endgraf
      \vspace{0.5em}\normalfont Adobe Search, Discovery \& ContentAI\endgraf
      \vspace{0.3em}{\small\ttfamily\{akadav,christanf,charora,vkotte,hongboz,lany,priyash,tking\}@adobe.com\endgraf}%
    }%
  }
  \hypersetup{%
    pdftitle={Softmax Reparameterization for Output-Head Quantization},
    pdfauthor={Asim Kadav, Christian Flores, Chirag Arora, Varun Kotte, Hongbo Zheng, Lan Yan, Priya Shanmugasundaram, Tracy Holloway King},
    hidelinks
  }
  \let\unblindedmaketitle\maketitle
  \renewcommand{\maketitle}{%
    \unblindedmaketitle\lhead{}%

  }
  \renewcommand{\headrulewidth}{0pt}
}

\usepackage[T1]{fontenc}
\usepackage{iclr2027_conference,times}
\usepackage{hyperref,url,graphicx,booktabs,amsmath,amssymb,microtype,enumitem}
\usepackage{tikz,pgfplots,algorithm,algpseudocode,longtable,array}
\usetikzlibrary{arrows.meta,positioning}
\pgfplotsset{compat=1.18}
\input{figure_style.tex}

\input{figdata/figure_numbers.tex}
\definecolor{paperblue}{RGB}{35,79,118}
\setlist{nosep,leftmargin=*}
\newcommand{\E}{\mathbb E}

\title{Softmax Reparameterization for Output-Head Quantization}
\begin{document}
\maketitle
\input{sections/abstract.tex}
\input{sections/introduction.tex}
\input{sections/method.tex}
\input{sections/experiments.tex}
\input{sections/analysis.tex}
\input{sections/related.tex}
\input{sections/conclusion.tex}
\input{sections/reproducibility.tex}
\bibliographystyle{iclr2027_conference}
\bibliography{references}
\clearpage
\appendix
\raggedbottom
\section*{Appendix}
\input{sections/appendix_guide.tex}
\input{sections/supp_base.tex}
\input{sections/app_nonlinear.tex}
\input{sections/app_robustness.tex}

\input{sections/app_stronger_controls.tex}
\input{sections/app_meancentering.tex}
\input{sections/app_mechanism.tex}
\input{sections/app_deployment.tex}
\input{sections/app_figdetails.tex}
\input{sections/supp_confidence.tex}
\input{sections/app_grouped.tex}
\end{document}

%% file: figure_style.tex
\usepgfplotslibrary{groupplots}
\usetikzlibrary{arrows.meta,calc,positioning,backgrounds,fit,decorations.pathreplacing}
\definecolor{figblue}{RGB}{0,114,178}
\definecolor{figorange}{RGB}{230,159,0}
\definecolor{figgreen}{RGB}{0,158,115}
\definecolor{figverm}{RGB}{213,94,0}
\definecolor{figpurple}{RGB}{117,112,179}
\definecolor{figgray}{RGB}{100,100,100}
\definecolor{figlight}{RGB}{225,225,225}
\pgfplotsset{
  paperaxis/.style={
    axis line style={black!55,line width=.4pt},
    tick style={black!45,line width=.4pt},
    tick label style={font=\scriptsize},
    label style={font=\scriptsize},
    title style={font=\small},
    grid=major,grid style={figlight,line width=.25pt},
    legend style={draw=none,fill=none,font=\scriptsize,cells={anchor=west}},
    every axis plot/.append style={line width=1.05pt},
  },
  bf16/.style={black,mark=*,mark size=1.7pt},
  minmax/.style={figgray,mark=square*,mark size=1.6pt},
  mse/.style={figorange,mark=triangle*,mark size=1.9pt},
  awmse/.style={figblue,mark=*,mark size=1.8pt},
  gptq/.style={figverm,mark=diamond,mark options={solid},mark size=2.1pt,densely dashed},
  awgptq/.style={figverm,mark=diamond*,mark size=2.1pt},
  codebook/.style={figgreen,mark=pentagon*,mark size=2pt,dashed},
}

%% file: sections/abstract.tex
\begin{abstract}
Large vocabularies make output heads a substantial inference cost in small
language models. We introduce \emph{softmax reparameterization}, a
post-training method that searches over functionally equivalent output heads
before quantization. The method subtracts a scalar multiple of the
vocabulary-row mean from every output row and selects the coefficient by
validation KL. For linear-softmax heads, these shifts preserve full-precision
predictions exactly and require no decoder retraining; a rank-one correction
extends the construction to nonlinear logit paths. Across seven output heads
and three quantizers, W4 gains are largest where baseline quantization
substantially distorts predictions: test KL falls by 93\% on XGLM under RTN
and by 73--77\% on Phi, BLOOM, and BLOOMZ under activation-weighted MSE.
Heads with low baseline error change little; at W2, used as a compression stress
test, benefits extend more broadly. On Phi, the gains persist under stronger
GPTQ calibration; a separate untouched holdout reproduces the improvements
on Phi and BLOOM. Frozen WikiText-selected coefficients also transfer without
retuning to C4 and OpenWebMath. Residual analysis on Phi shows how fidelity can improve despite
greater total logit error: the selected representative reduces error on likely
outputs and lowers its Fisher-weighted cost. For shift-compatible heads, the
shift adds no inference operation. With the decoder held in BF16, a packed W4
Phi output head reduces batch-one generation latency by 10.8\%, and
reparameterization preserves this speedup.
\end{abstract}

%% file: sections/introduction.tex
\section{Introduction}
\label{sec:introduction}

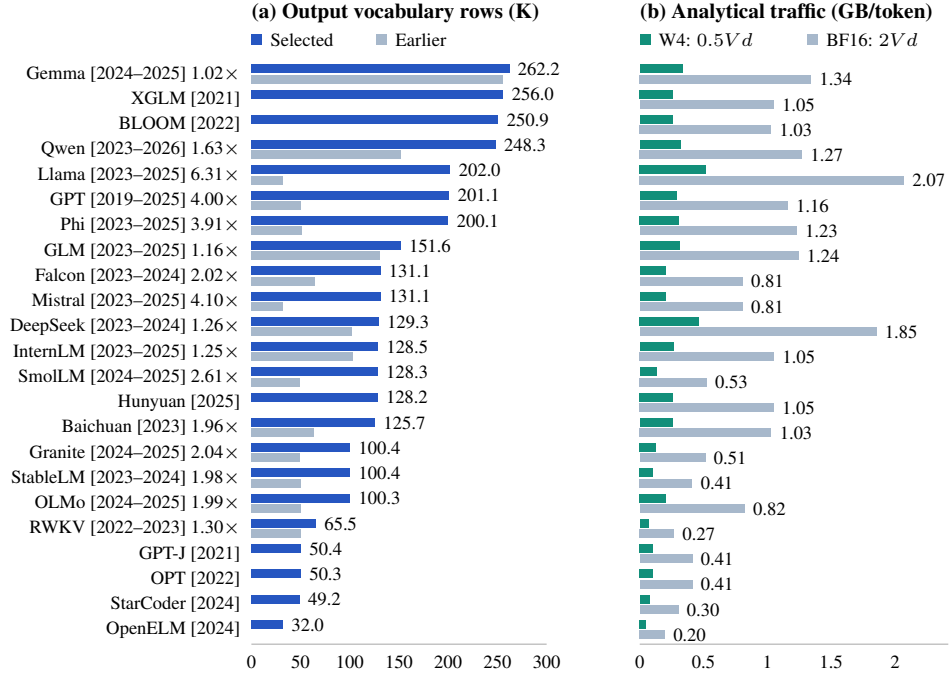
\begin{figure}[t]
\centering
\input{figures/vocabulary_expansion.tex}
\caption{\textbf{Large vocabularies create substantial output-head traffic.}
(a) Output rows across 23 selected model families, including small models such as Phi,
SmolLM, Granite and StableLM. Brackets give earlier--selected release years;
ratios compare selected with earlier vocabulary size. Pairs are selected examples,
not monotonic trajectories; missing earlier bars indicate no comparison.
(b) Analytical weight traffic for the selected checkpoint at batch-one decoding:
$2Vd$ bytes in BF16 versus $0.5Vd$ in W4, excluding scales and cache effects.
This is a storage-precision comparison, not a measurement of W4 quality or speed.}
\label{fig:head-size}
\end{figure}

Modern small language models (SLMs) pair modest decoder sizes with large multilingual vocabularies. For vocabulary size $V$ and hidden width $d$, the output projection contains $Vd$ weights, creating substantial per-token weight traffic (Figure~\ref{fig:head-size}). Unlike input embeddings, which retrieve selected rows, the dense output projection evaluates the full vocabulary at every decoding step \citep{grave2017efficientsoftmax}.

Vocabulary is itself a scaling dimension: scaling-law studies find that useful vocabulary capacity grows with model scale and that larger vocabularies can improve quality \citep{tao2024scalingvocab, huang2025overtokenized}. Figure~\ref{fig:head-size} shows that modern SLMs already deploy vocabularies far larger than historical 32K-scale designs, and each enlargement scales the head: the output projection performs about $Vd$ multiply-accumulates and reads about $2Vd$ bytes of weights per decoding step at bfloat16 (BF16) precision. Because single-token decoding is memory-bound, this weight movement is a direct cost, and low-bit kernels reduce it in proportion to the stored precision \citep{marlinartifact}, motivating output-head compression as vocabularies grow.

Despite this cost, practical post-training quantization recipes often retain the output head at higher precision \citep{llmcompressorimatrix}. The head directly determines next-token probabilities, without subsequent learned layers to absorb its error. In tied models, its source matrix also serves as the input embedding, introducing a separate deployment constraint \citep{kurtic2023sparsefinetuning}. Leaving the head unchanged therefore preserves a substantial per-token weight read even when the decoder is aggressively compressed.

Output-head quantization is often posed as preserving weights or logits, but
softmax depends only on relative logits. Equivalent full-precision output heads
can incur different quantization residuals (Figure~\ref{fig:reparam-equivalence}).
We exploit this freedom with
\emph{softmax reparameterization}, selecting an equivalent representative before
quantization.

Function-preserving reparameterizations are widely used in quantization,
including channel scaling and rotations
\citep{xiao2023smoothquant,quarot,spinquant}, while SageAttention applies fixed
mean-centering to attention keys before quantization
\citep{zhang2024sageattention}. Output Embedding Centering uses vocabulary-row
mean-centering during pretraining \citep{stollenwerk2026centering}.
Our contribution is a post-training scalar search within the output head's
additive softmax equivalence class, selecting the representative by
post-quantization predictive fidelity.

Let $\mu$ be the vocabulary-row mean of the output head. We search the scalar family
\[
W_t = W - t\mathbf1\mu^\top
\]
and select $t$ by the KL divergence of the quantized head to the source model on validation data. The search uses pretrained weights and requires no retraining or changes to the decoder. The search includes the original head ($t=0$) and ordinary mean-centering ($t=1$), and is performed separately for each base quantizer.

Across seven heads and three quantizers, gains are largest where baseline
quantization error is substantial, with broader benefits at W2
(Table~\ref{tab:reparam-breadth}). In packed Phi deployment, head compression
reduces batch-one and batch-16 latency by $10.8\%$ and $9.4\%$, respectively;
reparameterization preserves these speedups. Residual analysis shows how
fidelity can improve even when total logit error increases
(Section~\ref{sec:reparam-analysis}).

Our contributions are threefold:
\begin{itemize}[beginpenalty=10000]
\item \textbf{Softmax reparameterization.}
We introduce a post-training scalar search over exactly equivalent output-head parameterizations, selected by post-quantization prediction fidelity. A rank-one correction extends the construction to nonlinear logit paths.
\item \textbf{Low-bit recovery across models and quantizers.}
Across RTN, AW-MSE and full-Hessian GPTQ, W4 gains concentrate on substantially distorted heads; W2 benefits extend across nearly the full model--quantizer matrix. A separate untouched holdout reproduces the improvements on Phi and BLOOM.
\item \textbf{Deployment and mechanism.}
Reparameterization preserves the packed W4 latency benefit. Residual analysis shows that fidelity can improve despite greater reconstruction error when probability-weighted relative-logit error decreases.
\end{itemize}

%% file: figures/vocabulary_expansion.tex
\begin{tikzpicture}[x=1pt,y=1pt,every node/.style={font=\fontsize{7}{8}\selectfont,inner sep=0pt}]
\definecolor{vocabblue}{RGB}{38,86,193}
\definecolor{vocabgray}{RGB}{167,184,203}
\definecolor{vocabgreen}{RGB}{18,143,122}
\node[anchor=west,font=\fontsize{8}{9}\selectfont\bfseries] at (124,17) {(a) Output vocabulary rows (K)};
\node[anchor=west,font=\fontsize{8}{9}\selectfont\bfseries] at (270,17) {(b) Analytical traffic (GB/token)};
\fill[vocabblue] (124,5) rectangle (128,9);
\node[anchor=west] at (131,7) {Selected};
\fill[vocabgray] (171,5) rectangle (175,9);
\node[anchor=west] at (178,7) {Earlier};
\fill[vocabgreen] (270,5) rectangle (274,9);
\node[anchor=west] at (277,7) {W4: $0.5Vd$};
\fill[vocabgray] (333,5) rectangle (337,9);
\node[anchor=west] at (340,7) {BF16: $2Vd$};
\node[anchor=east] at (120,-6.000) {Gemma [2024--2025]  1.02$\times$};
\fill[vocabblue] (124,-5.650) rectangle (221.017,-2.550);
\fill[vocabgray] (124,-9.450) rectangle (218.720,-6.350);
\node[anchor=west] at (224.017,-4.100) {262.2};
\fill[vocabgreen] (270,-5.650) rectangle (286.110,-2.550);
\fill[vocabgray] (270,-9.450) rectangle (334.440,-6.350);
\node[anchor=west] at (337.440,-7.900) {1.34};
\node[anchor=east] at (120,-15.500) {XGLM [2021]};
\fill[vocabblue] (124,-15.150) rectangle (218.723,-12.050);
\node[anchor=west] at (221.723,-13.600) {256.0};
\fill[vocabgreen] (270,-15.150) rectangle (282.583,-12.050);
\fill[vocabgray] (270,-18.950) rectangle (320.333,-15.850);
\node[anchor=west] at (323.333,-17.400) {1.05};
\node[anchor=east] at (120,-25.000) {BLOOM [2022]};
\fill[vocabblue] (124,-24.650) rectangle (216.826,-21.550);
\node[anchor=west] at (219.826,-23.100) {250.9};
\fill[vocabgreen] (270,-24.650) rectangle (282.331,-21.550);
\fill[vocabgray] (270,-28.450) rectangle (319.325,-25.350);
\node[anchor=west] at (322.325,-26.900) {1.03};
\node[anchor=east] at (120,-34.500) {Qwen [2023--2026]  1.63$\times$};
\fill[vocabblue] (124,-34.150) rectangle (215.878,-31.050);
\fill[vocabgray] (124,-37.950) rectangle (180.216,-34.850);
\node[anchor=west] at (218.878,-32.600) {248.3};
\fill[vocabgreen] (270,-34.150) rectangle (285.257,-31.050);
\fill[vocabgray] (270,-37.950) rectangle (331.027,-34.850);
\node[anchor=west] at (334.027,-36.400) {1.27};
\node[anchor=east] at (120,-44.000) {Llama [2023--2025]  6.31$\times$};
\fill[vocabblue] (124,-43.650) rectangle (198.758,-40.550);
\fill[vocabgray] (124,-47.450) rectangle (135.840,-44.350);
\node[anchor=west] at (201.758,-42.100) {202.0};
\fill[vocabgreen] (270,-43.650) rectangle (294.828,-40.550);
\fill[vocabgray] (270,-47.450) rectangle (369.311,-44.350);
\node[anchor=west] at (372.311,-45.900) {2.07};
\node[anchor=east] at (120,-53.500) {GPT [2019--2025]  4.00$\times$};
\fill[vocabblue] (124,-53.150) rectangle (198.403,-50.050);
\fill[vocabgray] (124,-56.950) rectangle (142.595,-53.850);
\node[anchor=west] at (201.403,-51.600) {201.1};
\fill[vocabgreen] (270,-53.150) rectangle (283.899,-50.050);
\fill[vocabgray] (270,-56.950) rectangle (325.597,-53.850);
\node[anchor=west] at (328.597,-55.400) {1.16};
\node[anchor=east] at (120,-63.000) {Phi [2023--2025]  3.91$\times$};
\fill[vocabblue] (124,-62.650) rectangle (198.024,-59.550);
\fill[vocabgray] (124,-66.450) rectangle (142.944,-63.350);
\node[anchor=west] at (201.024,-61.100) {200.1};
\fill[vocabgreen] (270,-62.650) rectangle (284.750,-59.550);
\fill[vocabgray] (270,-66.450) rectangle (329.001,-63.350);
\node[anchor=west] at (332.001,-64.900) {1.23};
\node[anchor=east] at (120,-72.500) {GLM [2023--2025]  1.16$\times$};
\fill[vocabblue] (124,-72.150) rectangle (180.074,-69.050);
\fill[vocabgray] (124,-75.950) rectangle (172.295,-72.850);
\node[anchor=west] at (183.074,-70.600) {151.6};
\fill[vocabgreen] (270,-72.150) rectangle (284.898,-69.050);
\fill[vocabgray] (270,-75.950) rectangle (329.593,-72.850);
\node[anchor=west] at (332.593,-74.400) {1.24};
\node[anchor=east] at (120,-82.000) {Falcon [2023--2024]  2.02$\times$};
\fill[vocabblue] (124,-81.650) rectangle (172.497,-78.550);
\fill[vocabgray] (124,-85.450) rectangle (148.059,-82.350);
\node[anchor=west] at (175.497,-80.100) {131.1};
\fill[vocabgreen] (270,-81.650) rectangle (279.664,-78.550);
\fill[vocabgray] (270,-85.450) rectangle (308.655,-82.350);
\node[anchor=west] at (311.655,-83.900) {0.81};
\node[anchor=east] at (120,-91.500) {Mistral [2023--2025]  4.10$\times$};
\fill[vocabblue] (124,-91.150) rectangle (172.497,-88.050);
\fill[vocabgray] (124,-94.950) rectangle (135.840,-91.850);
\node[anchor=west] at (175.497,-89.600) {131.1};
\fill[vocabgreen] (270,-91.150) rectangle (279.664,-88.050);
\fill[vocabgray] (270,-94.950) rectangle (308.655,-91.850);
\node[anchor=west] at (311.655,-93.400) {0.81};
\node[anchor=east] at (120,-101.000) {DeepSeek [2023--2024]  1.26$\times$};
\fill[vocabblue] (124,-100.650) rectangle (171.834,-97.550);
\fill[vocabgray] (124,-104.450) rectangle (161.888,-101.350);
\node[anchor=west] at (174.834,-99.100) {129.3};
\fill[vocabgreen] (270,-100.650) rectangle (292.240,-97.550);
\fill[vocabgray] (270,-104.450) rectangle (358.961,-101.350);
\node[anchor=west] at (361.961,-102.900) {1.85};
\node[anchor=east] at (120,-110.500) {InternLM [2023--2025]  1.25$\times$};
\fill[vocabblue] (124,-110.150) rectangle (171.549,-107.050);
\fill[vocabgray] (124,-113.950) rectangle (162.172,-110.850);
\node[anchor=west] at (174.549,-108.600) {128.5};
\fill[vocabgreen] (270,-110.150) rectangle (282.633,-107.050);
\fill[vocabgray] (270,-113.950) rectangle (320.533,-110.850);
\node[anchor=west] at (323.533,-112.400) {1.05};
\node[anchor=east] at (120,-120.000) {SmolLM [2024--2025]  2.61$\times$};
\fill[vocabblue] (124,-119.650) rectangle (171.455,-116.550);
\fill[vocabgray] (124,-123.450) rectangle (142.186,-120.350);
\node[anchor=west] at (174.455,-118.100) {128.3};
\fill[vocabgreen] (270,-119.650) rectangle (276.304,-116.550);
\fill[vocabgray] (270,-123.450) rectangle (295.216,-120.350);
\node[anchor=west] at (298.216,-121.900) {0.53};
\node[anchor=east] at (120,-129.500) {Hunyuan [2025]};
\fill[vocabblue] (124,-129.150) rectangle (171.422,-126.050);
\node[anchor=west] at (174.422,-127.600) {128.2};
\fill[vocabgreen] (270,-129.150) rectangle (282.599,-126.050);
\fill[vocabgray] (270,-132.950) rectangle (320.397,-129.850);
\node[anchor=west] at (323.397,-131.400) {1.05};
\node[anchor=east] at (120,-139.000) {Baichuan [2023]  1.96$\times$};
\fill[vocabblue] (124,-138.650) rectangle (170.508,-135.550);
\fill[vocabgray] (124,-142.450) rectangle (147.680,-139.350);
\node[anchor=west] at (173.508,-137.100) {125.7};
\fill[vocabgreen] (270,-138.650) rectangle (282.356,-135.550);
\fill[vocabgray] (270,-142.450) rectangle (319.426,-139.350);
\node[anchor=west] at (322.426,-140.900) {1.03};
\node[anchor=east] at (120,-148.500) {Granite [2024--2025]  2.04$\times$};
\fill[vocabblue] (124,-148.150) rectangle (161.130,-145.050);
\fill[vocabgray] (124,-151.950) rectangle (142.186,-148.850);
\node[anchor=west] at (164.130,-146.600) {100.4};
\fill[vocabgreen] (270,-148.150) rectangle (276.166,-145.050);
\fill[vocabgray] (270,-151.950) rectangle (294.663,-148.850);
\node[anchor=west] at (297.663,-150.400) {0.51};
\node[anchor=east] at (120,-158.000) {StableLM [2023--2024]  1.98$\times$};
\fill[vocabblue] (124,-157.650) rectangle (161.130,-154.550);
\fill[vocabgray] (124,-161.450) rectangle (142.755,-158.350);
\node[anchor=west] at (164.130,-156.100) {100.4};
\fill[vocabgreen] (270,-157.650) rectangle (274.933,-154.550);
\fill[vocabgray] (270,-161.450) rectangle (289.730,-158.350);
\node[anchor=west] at (292.730,-159.900) {0.41};
\node[anchor=east] at (120,-167.500) {OLMo [2024--2025]  1.99$\times$};
\fill[vocabblue] (124,-167.150) rectangle (161.103,-164.050);
\fill[vocabgray] (124,-170.950) rectangle (142.612,-167.850);
\node[anchor=west] at (164.103,-165.600) {100.3};
\fill[vocabgreen] (270,-167.150) rectangle (279.858,-164.050);
\fill[vocabgray] (270,-170.950) rectangle (309.431,-167.850);
\node[anchor=west] at (312.431,-169.400) {0.82};
\node[anchor=east] at (120,-177.000) {RWKV [2022--2023]  1.30$\times$};
\fill[vocabblue] (124,-176.650) rectangle (148.248,-173.550);
\fill[vocabgray] (124,-180.450) rectangle (142.602,-177.350);
\node[anchor=west] at (151.248,-175.100) {65.5};
\fill[vocabgreen] (270,-176.650) rectangle (273.221,-173.550);
\fill[vocabgray] (270,-180.450) rectangle (282.885,-177.350);
\node[anchor=west] at (285.885,-178.900) {0.27};
\node[anchor=east] at (120,-186.500) {GPT-J [2021]};
\fill[vocabblue] (124,-186.150) rectangle (142.648,-183.050);
\node[anchor=west] at (145.648,-184.600) {50.4};
\fill[vocabgreen] (270,-186.150) rectangle (274.955,-183.050);
\fill[vocabgray] (270,-189.950) rectangle (289.818,-186.850);
\node[anchor=west] at (292.818,-188.400) {0.41};
\node[anchor=east] at (120,-196.000) {OPT [2022]};
\fill[vocabblue] (124,-195.650) rectangle (142.601,-192.550);
\node[anchor=west] at (145.601,-194.100) {50.3};
\fill[vocabgreen] (270,-195.650) rectangle (274.942,-192.550);
\fill[vocabgray] (270,-199.450) rectangle (289.768,-196.350);
\node[anchor=west] at (292.768,-197.900) {0.41};
\node[anchor=east] at (120,-205.500) {StarCoder [2024]};
\fill[vocabblue] (124,-205.150) rectangle (142.186,-202.050);
\node[anchor=west] at (145.186,-203.600) {49.2};
\fill[vocabgreen] (270,-205.150) rectangle (273.624,-202.050);
\fill[vocabgray] (270,-208.950) rectangle (284.496,-205.850);
\node[anchor=west] at (287.496,-207.400) {0.30};
\node[anchor=east] at (120,-215.000) {OpenELM [2024]};
\fill[vocabblue] (124,-214.650) rectangle (135.840,-211.550);
\node[anchor=west] at (138.840,-213.100) {32.0};
\fill[vocabgreen] (270,-214.650) rectangle (272.359,-211.550);
\fill[vocabgray] (270,-218.450) rectangle (279.437,-215.350);
\node[anchor=west] at (282.437,-216.900) {0.20};
\draw[black!35,line width=.3pt] (124,-220.500) -- (235.000,-220.500);
\node[anchor=north] at (124.000,-224.500) {0};
\node[anchor=north] at (142.500,-224.500) {50};
\node[anchor=north] at (161.000,-224.500) {100};
\node[anchor=north] at (179.500,-224.500) {150};
\node[anchor=north] at (198.000,-224.500) {200};
\node[anchor=north] at (216.500,-224.500) {250};
\node[anchor=north] at (235.000,-224.500) {300};
\draw[black!35,line width=.3pt] (270,-220.500) -- (386.000,-220.500);
\node[anchor=north] at (270.000,-224.500) {0};
\node[anchor=north] at (294.000,-224.500) {0.5};
\node[anchor=north] at (318.000,-224.500) {1};
\node[anchor=north] at (342.000,-224.500) {1.5};
\node[anchor=north] at (366.000,-224.500) {2};
\end{tikzpicture}

%% file: sections/method.tex
\section{Softmax Reparameterization}
\label{sec:reparam}

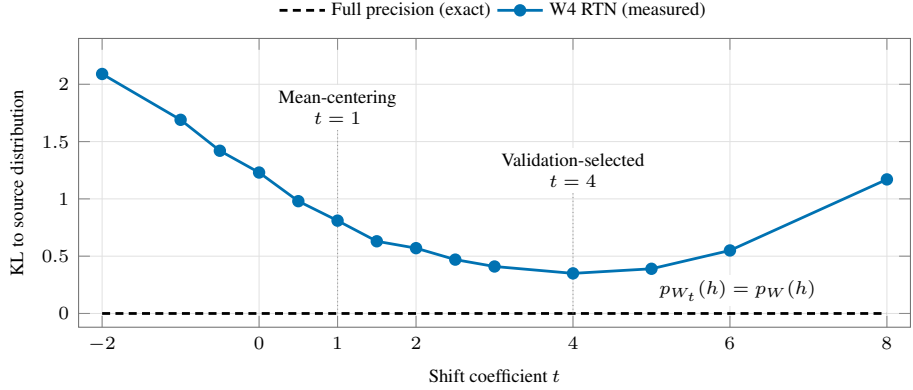
\begin{figure}[t]
\centering
\input{figures/reparam_equivalence.tex}
\caption{\textbf{Quantization fidelity across equivalent output-head parameterizations.}
Shared row shifts preserve predictions exactly (dashed: algebraic KL $=0$),
but change Phi-4-mini W4 RTN test KL (blue). Mean-centering uses $t=1$;
$t=4$ is selected on separate validation articles. All shifts use the same
quantizer and test states.}
\label{fig:reparam-equivalence}
\end{figure}

\subsection{Exact output-head equivalence}
The linear-softmax output head admits an exact additive reparameterization. Softmax depends only on relative logits, so subtracting
the same vector from every vocabulary row leaves the output distribution
unchanged. This gives a family of equivalent heads whose quantized fidelity can
differ.

Let $V$ be the vocabulary size and $d$ the hidden width. The output head
$W\in\mathbb R^{V\times d}$ maps a final hidden state $h\in\mathbb R^d$ to
logits $z=Wh$, giving next-token probabilities
$p_W(h)=\operatorname{softmax}(Wh)$. Let $\mathbf1\in\mathbb R^V$ denote the
all-ones vector. For any shared shift $a\in\mathbb R^d$, define
$W_a=W-\mathbf1a^\top$. Then
\begin{equation}
 W_a h=Wh-(a^\top h)\mathbf1,\qquad p_{W_a}(h)=p_W(h).
 \label{eq:reparam-invariance}
\end{equation}
The equality follows from cancellation of a common exponential factor in
softmax. It is an exact symmetry of a linear-softmax readout, independent
of the hidden-state distribution. An unchanged output bias also preserves
this identity. The shared row component is unidentifiable from prediction
probabilities.

Let $\mathcal B$ denote a fixed base quantization procedure, including its
precision, grouping, range objective and solver, and let $\mathcal B(W_a)$
denote its reconstructed weight matrix. In general, $\mathcal B(W_a)$ and
$\mathcal B(W)$ are not related by a vocabulary-wide shift. Define the
weight-quantization residual $E_a=\mathcal B(W_a)-W_a$.
Equation~\ref{eq:reparam-invariance} gives
\begin{equation}
 p_{\mathcal B(W_a)}(h)=\operatorname{softmax}(Wh+E_a h).
 \label{eq:reparam-error}
\end{equation}
The induced logit error is $E_a h$.
Reparameterization changes the quantization residual while preserving the
source distribution, but equivalence alone does not guarantee improved fidelity.

\subsection{Fidelity-selected reparameterization}
Low-bit quantization of the output head can substantially distort next-token
probabilities. To preserve output fidelity, we search over shared shifts of the
head weights before quantization and select the shift by validation KL.
Specifically, we restrict the $d$-dimensional shift to the vocabulary-row mean
direction:
\begin{equation}
 \mu=\frac1V W^\top\mathbf1,\qquad W_t=W-t\mathbf1\mu^\top,
 \qquad Q_t=\mathcal B(W_t;\mathcal H_{\rm fit}).
 \label{eq:reparam-family}
\end{equation}
The vocabulary-row mean identifies the shared component removed by ordinary
mean-centering. Varying its magnitude gives a low-cost, empirically effective
one-dimensional search within the exact equivalence class; we do not claim
that this path contains the optimal representative.
Figure~\ref{fig:reparam-equivalence} illustrates this search on Phi-4-mini:
full-precision predictions remain unchanged across shifts, while quantized KL
varies substantially.

The fitting states $\mathcal H_{\rm fit}$ are used only to fit the base
quantizer, while a disjoint set $\mathcal H_{\rm val}$ selects the shift.

For a finite grid $\mathcal T$ containing zero, the proposed selection rule is
\begin{equation}
 t^\star=\arg\min_{t\in\mathcal T} L_{\rm val}^{\mathcal B}(t),\qquad
 L_{\rm val}^{\mathcal B}(t)=\frac1{|\mathcal H_{\rm val}|}
 \sum_{h\in\mathcal H_{\rm val}}
 D_{\rm KL}\big(p_W(h)\,\Vert\,p_{Q_t}(h)\big).
 \label{eq:reparam-selection}
\end{equation}
We select by KL to preserve the source model's next-token distribution.
Perplexity can improve through a change in confidence even when predictions
depart further from the source (Appendix~\ref{app:confidence-controls}).
Final evaluation uses articles disjoint from quantizer fitting and shift selection.

Because $0\in\mathcal T$, exact evaluation of
Equation~\ref{eq:reparam-selection} gives
$L_{\rm val}^{\mathcal B}(t^\star)\leq L_{\rm val}^{\mathcal B}(0)$.
This guarantee applies only to validation KL under the quantizer used for
selection; it does not extend to unseen data or a different quantizer.
Appendix~\ref{app:base-quantizers} provides the coefficient grid,
quantizer-fitting details, and search pseudocode.

\subsection{Why fixed mean-centering is insufficient}

A natural choice is to subtract the vocabulary-row mean, corresponding to $t=1$.
This minimizes the full-precision weight norm, but does not necessarily preserve
predictions best after quantization. We therefore treat mean-centering as a
candidate in the search and allow validation KL to select a different shift.

Writing $W_c=W-\mathbf1\mu^\top$, the sum of its row vectors is zero
($\mathbf1^\top W_c=\mathbf0^\top$), so
\begin{equation}
 \|W_t\|_F^2=\|W_c\|_F^2+V(1-t)^2\|\mu\|_2^2.
 \label{eq:reparam-norm}
\end{equation}
Equation~\ref{eq:reparam-norm} makes precise what fixed mean-centering optimizes:
$t=1$ minimizes the full-precision Frobenius norm along this equivalence class.
Quantization instead depends on finite grids and rounding decisions
\citep{nagel2020adaround,zhang2025leanquant}. Changing $t$ changes groupwise
ranges, clipping, and rounding, and therefore changes the quantization residual.
The minimum-norm representative need not be the representative that best preserves
the predictive distribution after quantization. As an example, on Phi, validation KL selects $t=4$.

\paragraph{Applicability and inference.}
For a shift-compatible logit path, the selected shift is folded into the packed
head with no additional inference operation. Elementwise nonlinearities such as
tanh soft-capping require restoring $t(\mu^\top h)\mathbf1$ before the nonlinearity;
Appendix~\ref{app:reparam-nonlinear} gives the construction and equivalence check.
Tied models retain the source input embedding and quantize a separate output
copy \citep{kurtic2023sparsefinetuning}.

%% file: figures/reparam_equivalence.tex
\begin{tikzpicture}
\begin{axis}[
 paperaxis,width=.9\linewidth,height=5.4cm,
 xlabel={Shift coefficient $t$},ylabel={KL to source distribution},
 xmin=-2.3,xmax=8.3,ymin=-.12,ymax=2.4,
 xtick={-2,0,1,2,4,6,8},ytick={0,.5,1,1.5,2},
 legend style={at={(.5,1.03)},anchor=south,legend columns=2,
   font=\scriptsize,draw=none,fill=none},
]
\addplot[black,densely dashed,domain=-2:8,samples=2] {0};
\addlegendentry{Full precision (exact)}
\addplot[awmse] table[x=t,y=test_kl,col sep=comma]{figdata/reparam_curve.csv};
\addlegendentry{W4 RTN (measured)}
\draw[densely dotted,black!40] (axis cs:1,0)--(axis cs:1,1.6);
\draw[densely dotted,black!40] (axis cs:4,0)--(axis cs:4,1.05);
\node[font=\scriptsize,anchor=south,align=center,fill=white,inner sep=2pt]
 at (axis cs:1,1.6) {Mean-centering\\$t=1$};
\node[font=\scriptsize,anchor=south,align=center,fill=white,inner sep=2pt]
 at (axis cs:4,1.05) {Validation-selected\\$t=4$};
\node[font=\scriptsize,anchor=south,fill=white,inner sep=2pt]
 at (axis cs:6.1,.06) {$p_{W_t}(h)=p_W(h)$};
\end{axis}
\end{tikzpicture}

%% file: sections/experiments.tex
\section{Experiments}
\label{sec:reparam-experiments}

\paragraph{Experimental setup.}
We quantize only the output head while keeping the decoder fixed. Each comparison
measures how closely the quantized head reproduces its own source model's
predictions.

We evaluate seven heads: modern SLMs (Gemma 3/4, Qwen3.5 and Phi) and three
additional large-vocabulary models (BLOOM-1.7B, BLOOMZ-1.7B and XGLM-1.7B).
The W4 comparison uses G128 groups, BF16 decoders, and RTN, AW-MSE or full-Hessian
GPTQ \citep{frantar2023gptq} (Table~\ref{tab:reparam-breadth}). Before quantization, we verify
equivalence for every head: the linear-softmax heads reproduce
the source to $\mathrm{KL}\lesssim10^{-9}$, while Gemma~4 uses the rank-one
soft-cap correction of Equation~\ref{eq:reparam-rank1} (shifted BF16
KL $\approx7\times10^{-10}$, source PPL 66.41).

We use disjoint fitting, validation, and test splits: 128 WikiText \citep{wikitext} fitting
articles with eight states each, and 16 articles each for validation and test.
For each base quantizer, we select $t^\star$ from a 14-point grid over $[-2,8]$
by validation KL, then report source-to-candidate KL and PPL from FP32 readouts
on BF16-decoder states. Because the primary test split was available during
method development, we additionally evaluate the frozen method on a separate
untouched holdout.

KL values are token averages for fixed quantized heads. Key ablations include
article- or document-bootstrap confidence intervals; the BLOOM stability study
uses ten validation-subsampling seeds. These quantify evaluation-sample
uncertainty and coefficient-selection stability, respectively; they do not
measure sensitivity to alternative calibration samples or refitted quantizers.

\subsection{Output-head quantization results}
At W4, reparameterization recovers much of the fidelity lost on the most
distorted heads: test KL falls by $93\%$ on XGLM under RTN and by $73$--$77\%$
on Phi, BLOOM and BLOOMZ under AW-MSE (Table~\ref{tab:reparam-breadth}).
Full-Hessian GPTQ starts from lower baseline KL but still improves on these
heads. Gemma~3/4 and Qwen3.5 already have low W4 error and change little.
Across these comparisons, the largest gains occur where the base quantizer
leaves substantial error.

At W2, distortion rises across every evaluated head, and the benefit extends
across nearly the full model--quantizer matrix. Representative selection
therefore becomes useful more broadly as precision decreases, although
substantial KL reductions do not restore source-level fidelity in several W2
configurations.
We treat W2 as a compression stress test; Appendix~\ref{app:reparam-w23}
reports the full W3 and W2 results.

\begin{table}[t]
\centering\small
\caption{\textbf{Output-head quantization fidelity relative to the BF16 source model.}
Seven heads at W4 and W2 under RTN, AW-MSE and full-Hessian GPTQ.
Each cell reports test KL before ($t=0$) $\to$ after reparameterization
($t^\star$); lower is better. The shift is selected separately for each
quantizer by validation KL. Gemma~4 ($\dagger$) uses the rank-one soft-cap correction.}
\label{tab:reparam-breadth}
\resizebox{\textwidth}{!}{\input{generated/reparam_headline.tex}}
\end{table}

\subsection{Ablation studies and validation}
\label{sec:reparam-ablations}

\paragraph{Comparison with fixed mean-centering.}
Selecting the shift magnitude improves over fixed mean-centering on Phi,
BLOOM and BLOOMZ; on XGLM, all three quantizers select mean-centering itself
($t^\star=1$; Table~\ref{tab:reparam-centering}). On Phi, the validation-selected
$t=4$ lowers RTN test KL from about 0.81 at $t=1$ to 0.35, a further
$57\%$ reduction (Figure~\ref{fig:reparam-equivalence}).

Because the preferred representative also depends on the base quantizer, we
select $t$ separately for each quantizer throughout; transferring an RTN-selected
coefficient to AW-MSE can reduce fidelity (Appendix~\ref{app:reparam-matrix}).

\begin{table}[t]
\centering\small
\caption{\textbf{Fixed mean-centering versus validation-selected reparameterization at W4.}
Test KL relative to the BF16 source model for the unshifted head ($t=0$),
mean-centered head ($t=1$), and validation-selected head ($t^\star$), using
the Table~\ref{tab:reparam-breadth} evaluation split. Lower is better.}
\label{tab:reparam-centering}
\resizebox{\textwidth}{!}{\begin{tabular}{lccccccccc}
\toprule
 & \multicolumn{3}{c}{RTN} & \multicolumn{3}{c}{AW-MSE} & \multicolumn{3}{c}{GPTQ} \\
\cmidrule(lr){2-4}\cmidrule(lr){5-7}\cmidrule(lr){8-10}
Head & $t{=}0$ & $t{=}1$ & $t^\star$ & $t{=}0$ & $t{=}1$ & $t^\star$ & $t{=}0$ & $t{=}1$ & $t^\star$ \\
\midrule
Phi-4-mini & 1.229 & 0.811 & 0.351 & 0.936 & 0.577 & 0.256 & 0.158 & 0.117 & 0.059 \\
BLOOM-1.7B & 1.117 & 0.978 & 0.531 & 0.594 & 0.451 & 0.136 & 0.037 & 0.036 & 0.028 \\
BLOOMZ-1.7B & 1.119 & 0.859 & 0.483 & 0.662 & 0.399 & 0.158 & 0.042 & 0.039 & 0.032 \\
XGLM-1.7B & 2.129 & 0.143 & 0.143 & 0.584 & 0.095 & 0.095 & 0.009 & 0.007 & 0.007 \\
\bottomrule
\end{tabular}}
\end{table}

\begin{samepage}
\paragraph{Calibration, scaling, and affine quantization.}
\label{sec:stronger-controls}
On Phi, the benefit persists with more calibration data, exact channel
scaling, and affine quantization, indicating that it is not confined to the
original quantization setup. Increasing GPTQ calibration from 1,024 to 65,536
states, with the range initializer held fixed, lowers unshifted W4 test KL
from 0.160 to 0.089; reparameterization further reduces it to 0.035 ($61\%$).
Shifting also improves both RTN and AW-MSE after exact per-channel scaling,
and affine RTN with integer zero points. Paired article-bootstrap $95\%$
intervals exclude zero for each of these gains. Appendix~\ref{app:stronger-controls}
reports the full calibration, scaling, affine and direction controls.
\par
\end{samepage}

\paragraph{Evaluation on an independent holdout.}
We freeze the grid, selection rule, quantizers, and code, then evaluate once on
26 previously unused WikiText articles. AW-MSE KL falls by approximately $74\%$
on both Phi and BLOOM; RTN also improves, with paired article-bootstrap $95\%$
intervals excluding zero for all four reductions. Appendix~\ref{app:holdout}
reports the exact KL reductions, intervals, and likelihood measurements.

\paragraph{Cross-domain transfer and coefficient stability.}
We freeze the WikiText-selected coefficients and evaluate them without retuning
on C4 \citep{raffel2020t5} and OpenWebMath \citep{paster2024openwebmath}.
Across four heads, three W4 quantizers and two domains,
the frozen coefficient outperforms fixed mean-centering in all 18 comparisons
where the frozen WikiText-selected coefficient differs from $1$, and ties the
remaining six XGLM cases where it equals $1$.
In a separate BLOOM study, increasing validation size to 1,024 documents yields
identical selected coefficients across ten seeds for each C4/OpenWebMath--quantizer
pair; all selected coefficients remain different from $1$.
The preferred coefficient is nevertheless not universal: larger C4
and OpenWebMath validation sets can select different coefficients from WikiText
(Appendix~\ref{app:cross-domain-transfer}).

Appendix~\ref{app:grouped-extension} explores a higher-capacity group-wise
extension, with additional gains under W4 RTN and AW-MSE in a nine-model study
and a separate SmolLM3 evaluation with matched search budgets.

\subsection{Packed inference evaluation}
The preceding experiments evaluate quantized heads on captured hidden states.
We next test whether the fidelity gains persist in deployed generation and
measure the latency benefit of head compression. Here, packed W4 execution
stores weights as 4-bit codes and computes the output projection using the
Marlin INT4 kernel \citep{marlinartifact}.
We export zero-shift and $t=4$ Phi heads from the same BF16 source and evaluate
them through vLLM \citep{kwon2023vllm}, preserving all non-head tensors.

On 16,352 matched WikiText tokens, the shift lowers AW-MSE perplexity from 31.65
to 15.40 and KL to BF16 from 0.97 to 0.28 (Table~\ref{tab:reparam-deployment}).
It also improves full-Hessian GPTQ \citep{frantar2023gptq}, the strongest
deployed baseline evaluated here: perplexity falls from 13.67 to 12.23 and KL
from 0.15 to 0.06, leaving approximately $5\%$ perplexity excess over BF16's 11.65.
Thus representative selection adds fidelity even after GPTQ has compensated
rounding error, with the resulting head using the same INT4 format.

\begin{table}[t]
\centering\small
\caption{\textbf{Prediction fidelity and generation latency with packed W4 output heads on Phi-4-mini.}
Quality is evaluated on identical WikiText tokens. KL is measured relative to
BF16 using FP32 readout; perplexity uses vLLM-packed execution. A10G latency
covers 64-token greedy generation at batch sizes 1 and 16, with prefix caching
disabled. Dashes indicate unmeasured GPTQ latency.}
\label{tab:reparam-deployment}
\input{generated/reparam_deployment.tex}
\end{table}

The shift is folded into the weights before quantization, preserving the packed
layout and adding no inference operation for Phi's linear-softmax head.
Measured min--max and AW-MSE heads have essentially unchanged latency after
shifting; shifted AW-MSE has $10.8\%$ lower batch-one latency than BF16.
GPTQ uses the same Marlin format; its latency was not measured.

The tied Phi deployment retains its BF16 input embedding and adds a separate
packed output head, increasing resident model weights from 7.17 to 7.47\,GiB.
This memory cost applies equally to shifted and unshifted W4 heads.
Appendix~\ref{app:deployment-systems} gives the evaluation protocols and storage accounting.

%% file: generated/reparam_headline.tex
\begin{tabular}{lcccccc}
\toprule
 & \multicolumn{3}{c}{W4 KL} & \multicolumn{3}{c}{W2 KL} \\
\cmidrule(lr){2-4}\cmidrule(lr){5-7}
Model & RTN & AW-MSE & GPTQ & RTN & AW-MSE & GPTQ \\
\midrule
Phi-4-mini & 1.23$\to$0.351 & 0.936$\to$0.256 & 0.158$\to$0.059 & 98.2$\to$74.2 & 41.6$\to$16.2 & 3.08$\to$1.16 \\
Gemma 3 & 0.049$\to$0.046 & 0.041$\to$0.038 & 0.033$\to$0.033 & 2.66$\to$2.21 & 0.822$\to$0.714 & 0.508$\to$0.491 \\
Gemma 4$^\dagger$ & 0.012$\to$0.010 & 0.008$\to$0.008 & 0.008$\to$0.008 & 0.911$\to$0.911 & 0.212$\to$0.199 & 0.165$\to$0.162 \\
Qwen3.5 & 0.021$\to$0.019 & 0.013$\to$0.013 & 0.011$\to$0.010 & 1.84$\to$1.74 & 0.312$\to$0.285 & 0.200$\to$0.197 \\
BLOOM-1.7B & 1.12$\to$0.531 & 0.594$\to$0.136 & 0.037$\to$0.028 & 16.7$\to$6.15 & 4.46$\to$1.46 & 0.427$\to$0.335 \\
BLOOMZ-1.7B & 1.12$\to$0.483 & 0.662$\to$0.158 & 0.042$\to$0.032 & 16.2$\to$6.22 & 3.99$\to$2.12 & 0.456$\to$0.367 \\
XGLM-1.7B & 2.13$\to$0.143 & 0.584$\to$0.095 & 0.009$\to$0.007 & 97.8$\to$55.3 & 23.7$\to$3.73 & 0.173$\to$0.128 \\
\bottomrule
\end{tabular}

%% file: generated/reparam_deployment.tex
\begin{tabular}{lrrrr}
\toprule
Head & KL to BF16 & WikiText PPL & B1 (ms) & B16 (ms) \\
\midrule
BF16 & 0.00 & 11.65 & 1136.6 & 1335.8 \\
W4 min--max & 1.23 & 40.57 & 1015.3 & 1211.8 \\
W4 min--max + shift & 0.34 & 16.28 & 1014.1 & 1210.1 \\
W4 AW-MSE & 0.97 & 31.65 & 1014.8 & 1210.2 \\
W4 AW-MSE + shift & 0.28 & 15.40 & 1013.4 & 1210.3 \\
W4 GPTQ & 0.15 & 13.67 & --- & --- \\
W4 GPTQ + shift & 0.06 & 12.23 & --- & --- \\
\bottomrule
\end{tabular}

%% file: sections/analysis.tex
\section{Analysis of quantization error}
\label{sec:reparam-analysis}

We analyze how representative selection changes the quantization residual and
its effect on prediction fidelity. Reconstruction-based quantizers such as GPTQ
measure weight error through its effect on layer outputs
\citep{frantar2023gptq}, whereas KL also depends on the direction of the logit
perturbation relative to the source distribution. SoftWater similarly
incorporates feature covariance and softmax curvature into output-head
quantization \citep{cavalcanti2026softwater}.

\subsection{Softmax sensitivity to logit error}
We use the softmax Fisher matrix to quantify how logit errors affect the
predicted distribution. For small perturbations, its quadratic form approximates
KL divergence, accounting for both the magnitude and direction of the error
\citep{martens2020naturalgradient}.
This lets us examine why a shift can reduce KL even when total logit-error
energy increases.

For a fixed shift $t$, let $E=E_t=\mathcal B(W_t)-W_t$ be the weight residual
and $e=Eh$ the induced logit error. Write $p=p_W(h)$ for the source distribution
and $q=\operatorname{softmax}(Wh+e)$ for the perturbed distribution.
The forward KL, $D_{\rm KL}(p\Vert q)=\sum_{v=1}^V p_v\log(p_v/q_v)$,
weights each token's log-probability ratio by its source probability and is zero
when the distributions agree. For small $e$, it has the local expansion
\begin{equation}
 D_{\rm KL}(p\,\Vert\,\operatorname{softmax}(Wh+e))
 =\tfrac12 e^\top F(p)e+O(\|e\|_2^3),\qquad
 F(p)=\operatorname{diag}(p)-pp^\top,
 \label{eq:reparam-fisher}
\end{equation}
Here $F(p)$ is the Hessian of KL with respect to the logit perturbation at
$e=0$, and $\operatorname{diag}(p)$ places the source probabilities on the
diagonal. We call $\tfrac12 e^\top F(p)e$ the \emph{Fisher cost}; tables report
its mean over test states as Fisher/2.

\begin{samepage}
Its quadratic form is the source-probability-weighted variance of
the logit errors:
\[
 e^\top F(p)e=\sum_{v=1}^V p_v(e_v-\bar e_p)^2,
\qquad \bar e_p=\sum_{v=1}^V p_v e_v.
\]
\end{samepage}
Thus the approximation measures changes in relative logits, weighted by their
source probabilities, rather than total squared logit error. The
$O(\|e\|_2^3)$ remainder contains higher-order terms; the quadratic is a local
approximation whose accuracy we check against measured KL in Section~\ref{sec:empirical-residuals}.

This analysis applies directly to shift-compatible logit paths; for nonlinear
pre-softmax transformations such as Gemma~4's soft-cap, the Fisher quadratic
must use the post-transformation logit residual after the rank-one correction.
The categorical Fisher matrix satisfies $F(p)\mathbf1=0$, expressing the
common-shift invariance of softmax.
To separate this invisible component, let $P=I-\mathbf1\mathbf1^\top/V$ project
out the vocabulary-wide mean. All equivalent heads satisfy $PW_t=PW$, whereas
their quantized representatives need not satisfy $PQ_t=PQ_0$.

The hidden-state second moment $\Sigma=\E[hh^\top]$ determines logit-error
energy through $\E\|Eh\|_2^2=\operatorname{tr}(E\Sigma E^\top)$, while
$F(p)$ determines its local distributional cost. AW-MSE uses diagonal moments
of $\Sigma$ when fitting the quantizer. LeanQuant learns quantization grids
using inverse-Hessian error weights \citep{zhang2025leanquant}.
Here, the quantizer's objective remains fixed, and the Fisher quadratic diagnoses the
residuals produced by different representatives; selection uses actual
validation KL.

\subsection{Mechanism of fidelity recovery}
\label{sec:empirical-residuals}
On Phi-4-mini, the selected shift improves fidelity by reducing error where the
source model assigns probability mass. We compare fixed $t\in\{0,1,4\}$ on
$8{,}176$ test states disjoint from fitting and selection. Mean-centering
($t=1$) gives the smallest total logit error, but the validation-selected $t=4$
gives lower KL (Table~\ref{tab:reparam-mechanism}).

Softmax ignores offsets shared by all logits. This common-offset component is
larger at the selected $t=4$ than at $t=0$. Even after removing it, squared
error increases $2.8$-fold under RTN and $2.1$-fold under AW-MSE, while Fisher
cost falls by $72$--$73\%$. The improvement therefore depends on which relative
logits are affected, not just on removing a shared offset.

\begin{table}[t]
\centering\small
\setlength{\tabcolsep}{4pt}
\caption{\textbf{Lower KL despite greater logit error on Phi.}
Moving from $t=0$ to $t=4$ increases $D_z$ while reducing Fisher/2 and KL.
$D_z$ is source-normalized logit-error energy; Common is the fraction removed
by vocabulary centering. Fisher/2 denotes $\tfrac12\E[e^\top F(p)e]$.
Metrics use matched heads and states.}
\label{tab:reparam-mechanism}
\input{generated/reparam_mechanism.tex}
\end{table}

We group each token at each test state by its source probability. Bin $b$
contributes
$\tfrac12\E[\sum_{v\in b}p_v(e_v-\bar e_p)^2]$.
These contributions sum exactly to Fisher/2.
Under AW-MSE, the high- and middle-probability bins account for $71.8\%$ and
$27.9\%$ of its reduction, respectively
(Table~\ref{tab:reparam-fisher-attribution}).
Error more than doubles in the lowest-probability bin, yet that bin contributes
less than $0.5\%$ of Fisher cost at either shift. RTN gives almost the same
split: $73.2\%$ and $26.5\%$. Thus most of the extra reconstruction error falls
on outputs carrying little probability, while error on more probable outputs
decreases.

\begin{table}[t]
\centering\small
\setlength{\tabcolsep}{3pt}
\caption{\textbf{Probability-bin decomposition of the Fisher quadratic for Phi W4 AW-MSE.}
Means use the Table~\ref{tab:reparam-mechanism} test states. Visible logit error is
squared logit error after removing the vocabulary-wide common offset; Fisher/2 uses probability centering.
Mass is source probability mass; Drop is each bin's share of the total
Fisher/2 reduction from $t=0$ to $t=4$.}
\label{tab:reparam-fisher-attribution}
\input{generated/reparam_fisher_attribution.tex}
\end{table}

The quadratic differs from KL by at most $12.4\%$ across these comparisons
($1.4\%$ for RTN and $6.2\%$ for AW-MSE at $t=4$). The bins therefore explain
the reduction in a useful approximation to KL, rather than decomposing KL
itself. OSTQuant trains transformations with a KL-Top loss
\citep{hu2025ostquant}; we select frozen equivalent heads by full KL.

Full-Hessian GPTQ \citep{frantar2023gptq} shows the same error--fidelity tradeoff.
With $65{,}536$ calibration states, the selected $t=4$ raises logit-error energy
$2.66$-fold over $t=0$ but lowers Fisher cost and KL by $61\%$. On $8{,}176$
test states recaptured from the same articles, the quadratic matches KL within
$0.7\%$ at both shifts (Appendix~\ref{app:gptq-mechanism}). GPTQ fits logit
reconstruction without softmax curvature; representative selection can still
improve predictive fidelity.

%% file: generated/reparam_mechanism.tex
\begin{tabular}{llrrrrr}
\toprule
Quantizer & $t$ & $\|E_t\|/\|W_t\|$ & $D_z$ (\%) & Common (\%) & Fisher/2 & Actual KL \\
\midrule
RTN & 0 & 0.141 & 0.239 & 0.03 & 1.271 & 1.223 \\
RTN & 1 & 0.137 & 0.189 & 0.00 & 0.720 & 0.821 \\
RTN & 4 & 0.150 & 0.707 & 5.67 & 0.347 & 0.352 \\
AW-MSE & 0 & 0.125 & 0.249 & 4.06 & 0.861 & 0.935 \\
AW-MSE & 1 & 0.121 & 0.191 & 0.13 & 0.532 & 0.583 \\
AW-MSE & 4 & 0.136 & 0.677 & 27.51 & 0.240 & 0.256 \\
\bottomrule
\end{tabular}

%% file: generated/reparam_fisher_attribution.tex
\begin{tabular}{lrrrr}
\toprule
Source probability & Mass (\%) & Visible logit error, $t=0\to4$ & Fisher/2, $0\to4$ & Drop (\%) \\
\midrule
$p\geq10^{-2}$ & 84.766 & $35.00 \to 13.48$ & $0.60557 \to 0.15943$ & 71.8 \\
$10^{-6}\leq p<10^{-2}$ & 15.037 & $10,812.73 \to 5,279.47$ & $0.25264 \to 0.07930$ & 27.9 \\
$p<10^{-6}$ & 0.196 & $132,569.58 \to 288,852.93$ & $0.00300 \to 0.00116$ & 0.3 \\
\bottomrule
\end{tabular}

%% file: sections/related.tex
\section{Related Work}
\label{sec:related}

\paragraph{Quantization and equivalent parameterizations.}
SmoothQuant redistributes activation and weight scales, AWQ uses activation
statistics to choose channel scales, and QuaRot and SpinQuant use rotations
to improve low-precision inference
\citep{xiao2023smoothquant,lin2024awq,quarot,spinquant}.
These methods exploit function-preserving changes in parameterization
to improve quantization. Concurrent GaugeQuant learns quantization-friendly
bases from internal Transformer symmetries during training
\citep{gaugequant2026}; our main method instead operates post-training on the
additive symmetry specific to the output head, searches a scalar family,
and selects by the prediction fidelity of the quantized frozen model.

\paragraph{Softmax invariance and output centering.}
SageAttention applies fixed mean-centering to sequence-dependent key activations
\citep{zhang2024sageattention}; we instead select the magnitude of a static
output-weight shift according to post-quantization predictive fidelity.
Output Embedding Centering uses the
vocabulary-row mean to stabilize language-model pretraining
\citep{stollenwerk2026centering}. Consequently, neither the softmax symmetry
nor mean-centering is a new contribution here.
Table~\ref{tab:reparam-centering} quantifies how much the selected amount adds
over fixed centering across heads and quantizers.

\paragraph{Reconstruction and distributional objectives.}
AdaRound, BRECQ, Optimal Brain Compression and GPTQ develop reconstruction
objectives and compensation procedures for post-training quantization
\citep{nagel2020adaround,li2021brecq,frantar2022obc,frantar2023gptq}.
Importance-matrix implementations provide diagonal activation-weighted
fitting \citep{llamacppimatrix,llmcompressorimatrix}. Concurrent SoftWater uses feature
covariance and softmax curvature for class-aware quantization rate allocation
\citep{cavalcanti2026softwater}. We use curvature as a diagnostic and measured
KL to select a representative before applying a base quantizer.

\paragraph{Large-vocabulary prediction and execution.}
Adaptive softmax changes the organization of large-vocabulary prediction
\citep{grave2017efficientsoftmax}. ARCHead compresses output heads with low-rank
structure and an INT4 residual correction \citep{archead2026}; our intervention
instead keeps the dense head and only reparameterizes it before an existing
quantizer. Dense head quantization retains all vocabulary outputs. Tied-weight
separation and packed Marlin/vLLM
execution are established deployment tools
\citep{kurtic2023sparsefinetuning,marlinartifact,kwon2023vllm}; our serving
measurements quantify their relevance to output-head compression.

%% file: sections/conclusion.tex
\section{Conclusion}
Functionally equivalent output heads can exhibit markedly different low-bit behavior. At W4, gains concentrate on heads with substantial baseline quantization error; at W2, where distortion increases across all evaluated heads, the benefit broadens across nearly the full model--quantizer matrix. The gains persist across three quantizers and stronger calibration, scaling and affine-quantization controls. A separate untouched holdout confirms the improvements on Phi and BLOOM. Matched residual analysis shows that a better representative need not reduce reconstruction error, but can place that error in directions that matter less to the predictive distribution. Packed inference confirms that this fidelity recovery preserves the W4 latency benefit. More broadly, an output head's apparent precision requirement can depend on its parameterization, not only on the function it represents.

%% file: sections/reproducibility.tex
\section*{Reproducibility Statement}
Appendix~\ref{app:base-quantizers} specifies data splits, the search grid,
quantizers, and numerical conventions; Appendix~\ref{app:checkpoints} lists
checkpoints and head dimensions. Appendices~\ref{app:robustness}--\ref{app:deployment-systems}
and~\ref{app:grouped-extension} document evaluation protocols, sample counts,
uncertainty estimates, seeds, hardware, and timing procedures. Upon publication,
we will release quantization and evaluation code, machine-readable experiment
records, and scripts for regenerating the principal result tables.

%% file: sections/appendix_guide.tex
Appendices~\ref{app:base-quantizers} and \ref{app:reparam-nonlinear} give the
experimental protocol, search pseudocode, base quantizers, and nonlinear extension.
Appendices~\ref{app:robustness}--\ref{app:mechanism} report robustness and transfer
results, stronger quantization controls, mean-centering comparisons, and residual
diagnostics. Appendices~\ref{app:deployment-systems}--\ref{app:grouped-extension}
cover deployment, model and vocabulary details, confidence controls, and grouped
search. Numerical paths and BF16 references are matched within each comparison;
results from different protocols are not pooled.

%% file: sections/supp_base.tex
\section{Experimental Protocol and Base Quantizers}
\label{app:base-quantizers}
\label{sec:method}

\subsection{Data, selection and evaluation}
\label{app:selection-protocol}
The matrix and breadth probes capture final hidden states from frozen BF16
models. A fixed WikiText partition assigns 128 articles to quantizer fitting,
with eight positions per article (1,024 states), and 16 articles each to
validation and test. Prefixes contain at most 512 tokens. Fitting estimates
activation moments; validation selects the coefficient; test reports KL, top-1
agreement and perplexity. Linear-logit products and scoring use FP32. Gemma~4 instead applies
its soft-cap after the correction in Appendix~\ref{app:reparam-nonlinear}.

The ordered candidate grid is
\[
 \mathcal T=(-2,-1,-0.5,0,0.5,1,1.5,2,2.5,3,4,5,6,8).
\]
The grid includes the original head ($t=0$) and fixed mean-centering ($t=1$),
with finer spacing near these baselines, coarser spacing at larger positive
shifts, and negative candidates to test the opposite direction. The same grid
is used for every model and base quantizer in the matrix and breadth evaluations.
Each base quantizer selects its own coefficient from unrounded validation KL.
The recompute implementations retain the first candidate on an exact tie.
They reuse fitting moments across candidates: AW-MSE uses diagonal moments,
and GPTQ reuses the same fitting Hessian while quantizing each representative
independently. The four-head RTN/AW-MSE search takes about five minutes on one
A100, and Phi's 14-point GPTQ sweep about twelve.

The separate 26-article untouched holdout
is reported in Appendix~\ref{app:holdout}. A plotted evaluation curve is
post-selection analysis; it does not provide a new selection set. Small changes
in the diagnostic matrix lack per-article uncertainty. Packed deployment uses a
separate block-based protocol and source reference (Appendix~\ref{app:reparam-integrated}).

\begin{algorithm}[H]
\caption{Validation-selected softmax reparameterization}
\label{alg:reparam-search}
\small
\begin{algorithmic}[1]
\Require Source head $W$, logit path $g$, base quantizer $\mathcal B$;
\Statex \hspace{\algorithmicindent} fitting states $\mathcal H_{\rm fit}$, validation states $\mathcal H_{\rm val}$;
\Statex \hspace{\algorithmicindent} fixed ordered grid $\mathcal T$ containing $0$
\State $\mu\gets W^\top\mathbf1/V$; estimate quantizer moments from $\mathcal H_{\rm fit}$
\State Set $r=0$ if $g$ preserves common shifts, otherwise $r=1$
\State Verify unquantized equivalence for all $t\in\mathcal T$, correcting if $r=1$
\State $L^\star\gets+\infty$
\For{$t$ in $\mathcal T$, in order}
  \State $W_t\gets W-t\mathbf1\mu^\top$; $Q_t\gets\mathcal B(W_t;\mathcal H_{\rm fit})$
  \State $p(h)\gets\operatorname{softmax}(g(Wh))$
  \State $q_t(h)\gets\operatorname{softmax}(g(Q_t h+r\,t(\mu^\top h)\mathbf1))$
  \State $L_t\gets |\mathcal H_{\rm val}|^{-1}\sum_{h\in\mathcal H_{\rm val}}D_{\rm KL}(p(h)\Vert q_t(h))$
  \If{$L_t<L^\star$}
    \State $(L^\star,t^\star,Q^\star)\gets(L_t,t,Q_t)$
  \EndIf
\EndFor
\State \Return $Q^\star$; retain $(t^\star,\mu)$ for inference only if $r=1$
\end{algorithmic}
\end{algorithm}

Algorithm~\ref{alg:reparam-search} uses the actual model logit path; an unchanged
bias can be included in $g$. Final evaluation states are not inputs to the
search. Quantizer precision, groups, clipping grid and numerical conventions
remain fixed across candidates. The equivalence check permits numerical
roundoff.

\subsection{Base quantizers}
Let $k_b=2^{b-1}-1$. For a row group, the candidate scale is
$s_g(c)=c\max_{j\in g}|W_{vj}|/k_b$, with a positive numerical floor for zero
ranges. Codes round $W_{vj}/s_g(c)$ and clip to the integer range specified in
Table~\ref{tab:numerical-conventions}. The table also specifies whether scales
are cast to BF16 before candidate scoring or only when stored.

RTN fixes $c=1$. AW-MSE uses $m_j=\E_{\rm fit}[h_j^2]$ and selects
\begin{equation}
 c_g^\star=\arg\min_{c\in\mathcal C}\sum_{j\in g}m_j\big(W_{vj}-Q_{vj}(c)\big)^2.
 \label{eq:scale}
\end{equation}
This is an established importance-weighted primitive
\citep{llamacppimatrix,llmcompressorimatrix}; unweighted MSE sets $m_j=1$.
GPTQ uses AW-MSE initial scales and propagates rounding error with the full
fitting second moment $H=2X^\top X/N$, 1\% damping, block size 128 and no
activation ordering \citep{frantar2023gptq}. All heads use G128 unless noted.

\subsection{Numerical conventions by experiment}
\label{app:numerical-conventions}
For the matrix AW-MSE/GPTQ and matched probability probes, the clipping grid is
\[
 \mathcal C=(1,.975,.95,.925,.9,.875,.85,.8,.75,.7,.6,.5).
\]
The packed Phi AW-MSE path uses the first ten factors, ending at $0.7$.
RTN uses only $c=1$. A raw/shifted pair always shares its convention;
cross-quantizer comparisons can also differ in scale precision and integer range.

\begin{table}[ht]
\centering\footnotesize
\setlength{\tabcolsep}{3pt}
\caption{Numerical conventions for the reported comparisons.
``Signed'' is $[-2^{b-1},k_b]$; ``symmetric'' is $[-k_b,k_b]$.
Reconstruction precision describes the weights presented to the scoring path.}
\label{tab:numerical-conventions}
\begin{tabular}{@{}>{\raggedright\arraybackslash}p{.27\linewidth}>{\raggedright\arraybackslash}p{.14\linewidth}>{\raggedright\arraybackslash}p{.32\linewidth}>{\raggedright\arraybackslash}p{.19\linewidth}@{}}
\toprule
Comparison & Codes & Scale selection/storage & Reconstruction \\
\midrule
Matrix/breadth RTN & Symmetric & FP32 & FP32 \\
Matrix/breadth AW-MSE & Signed & Score and store BF16 & BF16, then FP32 products \\
GPTQ probe & Signed & BF16 initial scales & BF16, then FP32 products \\
Matched Phi residuals & Signed & Score and store BF16 & BF16, then FP32 products \\
Packed Phi RTN/AW-MSE & Signed & Float candidates; store BF16 & Packed W4 serving \\
\bottomrule
\end{tabular}
\end{table}

The reconstruction scan in Appendix~\ref{app:reparam-scan} is a separate
symmetric-RTN diagnostic with its own shorter coefficient grid. Integrated
likelihood measurements use the serving model's logit path and BF16 reference;
probability probes evaluate the specified logit transform and softmax directly,
without deployment sampling penalties.

\paragraph{Implementation and evidence provenance.}
\label{app:implementation}
Packing, tied-weight separation and kernel verification are documented in
Appendix~\ref{app:deployment-systems}. The measured packed Phi path is
shift-compatible and requires no rank-one correction. Gemma~4's correction is
evaluated in the dense probability probes; its packed deployment was not measured.
The build script regenerates the matrix, breadth and transfer tables
from per-cell records, including record hashes, selected coefficients and the
transfer-regression check. Captured-state hashes and checkpoint revisions are
also needed for bit-exact reproduction.

%% file: sections/app_nonlinear.tex
\section{Nonlinear Logit Paths and Inference Compatibility}
\label{app:reparam-nonlinear}

Some models apply a nonlinear transformation to logits before softmax. A common
shift before that transformation can become unequal changes afterward, altering
the predictions. For these models, we restore the removed shared component
before applying the nonlinearity.

\subsection{Shift-compatible logit paths}
Equation~\ref{eq:reparam-invariance} assumes the complete pre-softmax logit path
preserves vocabulary-wide additive shifts. Let $g$ denote every operation applied
to the linear logits before softmax. The reparameterization is exact precisely
when
\begin{equation}
 \operatorname{softmax}\big(g(z+c\mathbf1)\big)=\operatorname{softmax}\big(g(z)\big)
 \quad\text{for all }z,c,
 \label{eq:reparam-applicability}
\end{equation}
Equivalently, $P\,g(z+c\mathbf1)=P\,g(z)$ with
$P=I-\mathbf1\mathbf1^\top/V$. Identity logits, temperature scaling and an
unchanged additive bias satisfy this; an elementwise nonlinearity such as
tanh logit soft-capping does not, because $s\tanh((z+c\mathbf1)/s)$ is not a common
shift of $s\tanh(z/s)$.

\subsection{Rank-one correction}
When $g$ violates Equation~\ref{eq:reparam-applicability}, the shared component can
be retained exactly rather than discarded. Decompose the head as
$W=(W-t\mathbf1\mu^\top)+t\mathbf1\mu^\top$, quantize only the first term to
$Q_t$, and restore the rank-one term to the quantized logits \emph{before} $g$:
\begin{equation}
 z_t=Q_t h+t(\mu^\top h)\mathbf1,\qquad p_t=\operatorname{softmax}\big(g(z_t)\big).
 \label{eq:reparam-rank1}
\end{equation}
With $Q_t=W_t$ this gives $z_t=Wh$ exactly, so $p_t=p_W$ for \emph{any} $g$; on
Gemma-4 the unquantized shifted head reproduces the source distribution to
$\text{KL}\approx7\times10^{-10}$. After quantization, the only residual entering
$g$ is $E_t h$, where $E_t=Q_t-W_t$. The correction
costs one $d$-dimensional dot product and a scalar broadcast over the vocabulary,
compared with $Vd$ multiply-accumulates for the projection, and can be fused
into a soft-capping pass; for a pure softmax path it may be omitted because softmax discards
$\mathbf1$. We still execute each shifted head through the model's actual logit
path and verify $p_t=p_W$ before searching $t$, since casting to BF16 introduces
roundoff and the check is empirical rather than a claim of bitwise equivalence.

\paragraph{Storage and tied weights.}
The correction retains $t$ and the $d$-dimensional row mean $\mu$, an $O(d)$
overhead relative to the $O(Vd)$ head. For tied weights, the input embedding
retains its source values while a separate output copy is quantized, using
established weight separation \citep{kurtic2023sparsefinetuning}.
Appendix~\ref{app:deployment} gives the resulting storage accounting.

The Gemma~4 equivalence and fidelity measurements use dense probability probes.
The packed timing results in Appendix~\ref{app:reparam-integrated} concern Phi
and do not measure the runtime cost of this correction.

%% file: sections/app_robustness.tex
\section{Extended Robustness and Generalization}
\label{app:robustness}

This section reports an untouched-data replication in two independent model
families, multilingual and cross-domain transfer of frozen English-selected
coefficients, validation-size stability, and the extension to lower precisions.

\subsection{Untouched holdout replication}
\label{app:holdout}
Section~\ref{sec:reparam-experiments} reports the Phi and BLOOM results on the
26-article untouched holdout (13,286 tokens). Article-bootstrap 95\% intervals
use 10,000 replicates: Phi KL reductions are $[72.9,74.4]\%$ under AW-MSE and
$[70.4,71.7]\%$ under RTN; BLOOM intervals are $[72.4,76.2]\%$ and
$[48.7,54.3]\%$, respectively. The per-cell records accompany the code release.
Re-running the frozen rule on the main test split reproduces the Phi
matrix values exactly (RTN 1.229 to 0.351 KL, AW-MSE 0.936 to 0.256;
AW-MSE PPL 24.72 to 12.50, BF16 9.73). This checks implementation consistency;
the separate holdout supplies the generalization test.

Table~\ref{tab:holdout-details} retains the likelihood measurements and exact KL
pairs. BLOOM GPTQ improves by 23.4\% using unrounded KL values; no bootstrap
interval is reported for that comparison.

\begin{table}[ht]
\centering\small
\caption{\textbf{KL reductions persist on untouched holdouts.} Pairs are original $\to$ shifted;
BF16 PPL uses the same source and evaluation path. Dashes denote unreported values.}
\label{tab:holdout-details}
\begin{tabular}{llrrr}
\toprule
Head & Quantizer & KL & PPL & BF16 PPL \\
\midrule
Phi & AW-MSE & $0.943\to0.249$ & $23.41\to11.93$ & 9.33 \\
Phi & RTN & $1.204\to0.348$ & $31.00\to12.93$ & 9.33 \\
BLOOM & AW-MSE & $0.558\to0.142$ & --- & --- \\
BLOOM & GPTQ & $0.037\to0.028$ & --- & --- \\
\bottomrule
\end{tabular}
\end{table}

\begin{samepage}
\subsection{Frozen English-selected shifts on FLORES}
\label{app:flores-transfer}
We evaluate the existing English WikiText-selected W4 G128 heads on the first
256 aligned sentences of the original FLORES-200 devtest release, independently
in each of six languages, without refitting or retuning $t$.\footnote{\href{https://github.com/facebookresearch/flores/blob/main/flores200/README.md}{Meta's original FLORES-200 release.}}
The BF16 decoder is frozen; each sentence supplies at most 128 tokens (none
were truncated), and Table~\ref{tab:flores-transfer} reports token-weighted,
full-vocabulary $D_{\rm KL}(p_W\Vert p_Q)$ on all next-token positions, using
FP32 readouts and FP64 reductions with TF32 disabled. XGLM improves in all
18 language--quantizer point estimates, with 17 paired intervals excluding
zero. BLOOM instead increases KL by 67.3\% for Arabic/RTN, 13.4\% for
Hindi/RTN, and 82.7\% for Hindi/AW-MSE; all three intervals exclude zero.
Its small Chinese/GPTQ increase is uncertain. Both models improve under all
three quantizers on English FLORES. These results expose distribution-dependent
transfer of frozen heads; they neither guarantee multilingual transfer nor
establish that the residual moved specifically into non-English token coordinates.
The FLORES sentences here probe short-context next-token fidelity rather than translation quality.

\end{samepage}

\begin{table}[htbp]
\centering\small
\setlength{\tabcolsep}{3pt}
\caption{\textbf{English-selected shifts lower all 18 XGLM FLORES KL values but can regress on BLOOM.}
Entries are raw $\to$ shifted KL (nats). BLOOM uses $t=(6,4,4)$ for
(RTN, AW-MSE, GPTQ); XGLM uses $t=1$. GPTQ retains the original
1,024-state Hessian and ranges. Bold marks increased KL; $\dagger$ marks
paired 95\% intervals containing zero: 5,000 bootstrap resamples of 75
source-article URL clusters, without multiplicity adjustment.
Arabic is Modern Standard Arabic; Chinese uses Simplified script.}
\label{tab:flores-transfer}
\input{generated/flores_transfer.tex}
\end{table}

\input{sections/app_cross_domain.tex}

\subsection{Lower-precision extension (W3 and W2)}
\label{app:reparam-w23}
Table~\ref{tab:reparam-w23} extends the seven-head comparison to W3 and W2 using
the protocol in Appendix~\ref{app:selection-protocol}. At W3, Phi test KL falls
from 12.97 to 2.46 under RTN, 4.62 to 1.12 under AW-MSE, and 0.543 to 0.211
under GPTQ. W2 remains substantially degraded even with the shift: Phi GPTQ
reaches PPL 30.0 versus the matched BF16 reference of 9.73, while Gemma~3 RTN
falls from 248.3 to 146.3. The relative KL reduction is largest for RTN and
smallest for GPTQ at W3; this ordering does not persist at W2. No head's test
KL increases in this comparison.

\begin{table}[ht]
\centering\small
\caption{\textbf{Shifting lowers or preserves KL at W3 and W2, but W2 distortion remains large.}
Each cell is raw ($t=0$) $\to$ reparameterized ($t^\star$) test KL to the source,
with per-quantizer validation selection over the same 14-point grid.
The four main heads appear above each rule; three very-large-vocabulary
heads appear below. Gemma~4 ($\dagger$) uses the rank-one soft-cap correction.}
\label{tab:reparam-w23}
\input{generated/reparam_w23.tex}
\end{table}

Table~\ref{tab:reparam-w23-ppl} gives W3 perplexities from FP32 readouts on the
same 8,176-position evaluation split. Its BF16 column is the unquantized source
head on the same states. Very large RTN values reflect near-degenerate heads:
Phi falls from $3.98\times10^6$ to 111.5 with the shift, and XGLM from
$1.5\times10^{14}$ to 34.5 (BF16 13.8). Phi GPTQ reaches 11.97, compared with
BF16 9.73.

\begin{table}[ht]
\centering\small
\caption{\textbf{KL-selected shifts can substantially reduce W3 perplexity, but gains are not uniform.}
Each quantizer cell is raw ($t=0$) $\to$ reparameterized ($t^\star$) perplexity.
BF16 is the unquantized source head on the same 8{,}176-position test split
(FP32 readout). Gemma~4 ($\dagger$) uses the rank-one soft-cap correction.}
\label{tab:reparam-w23-ppl}
\input{generated/reparam_w23_ppl.tex}
\end{table}

%% file: generated/flores_transfer.tex
\begin{tabular}{@{}llrrr@{}}
\toprule
Model & Language & RTN & AW-MSE & GPTQ \\
\midrule
BLOOM-1.7B & English & $1.586599\!\to\!0.665851$ & $0.920758\!\to\!0.144382$ & $0.055268\!\to\!0.038390$ \\
 & French & $0.946322\!\to\!0.556638$ & $0.845735\!\to\!0.208255$ & $0.713031\!\to\!0.250820$ \\
 & Spanish & $2.074753\!\to\!0.762232$ & $1.565073\!\to\!0.273351$ & $0.815738\!\to\!0.297205$ \\
 & Arabic & $\mathbf{0.523881}\!\to\!\mathbf{0.876482}$ & $0.150436\!\to\!0.148087^{\dagger}$ & $0.236476\!\to\!0.209441$ \\
 & Hindi & $\mathbf{1.629840}\!\to\!\mathbf{1.847837}$ & $\mathbf{0.230221}\!\to\!\mathbf{0.420559}$ & $0.951959\!\to\!0.447360$ \\
 & Chinese & $0.641800\!\to\!0.625781$ & $0.083138\!\to\!0.067635$ & $\mathbf{0.118631}\!\to\!\mathbf{0.119211}^{\dagger}$ \\
\midrule
XGLM-1.7B & English & $1.849360\!\to\!0.167656$ & $0.427737\!\to\!0.104735$ & $0.011856\!\to\!0.008691$ \\
 & French & $1.250451\!\to\!0.168155$ & $0.214890\!\to\!0.106347$ & $0.022430\!\to\!0.017270$ \\
 & Spanish & $1.040484\!\to\!0.169850$ & $0.229209\!\to\!0.117290$ & $0.023801\!\to\!0.019992$ \\
 & Arabic & $0.658988\!\to\!0.235897$ & $0.189734\!\to\!0.171468$ & $0.052101\!\to\!0.051720^{\dagger}$ \\
 & Hindi & $1.249754\!\to\!0.254091$ & $0.263036\!\to\!0.163331$ & $0.058555\!\to\!0.048005$ \\
 & Chinese & $1.610301\!\to\!0.227931$ & $0.312039\!\to\!0.155611$ & $0.031372\!\to\!0.029112$ \\
\bottomrule
\end{tabular}

%% file: sections/app_cross_domain.tex
\subsection{Cross-domain transfer and coefficient stability}
\label{app:cross-domain-transfer}

\paragraph{Frozen WikiText coefficients.}
We evaluate four heads under W4 G128 RTN, AW-MSE and full-Hessian GPTQ on
C4 English and OpenWebMath. Let $t_{\rm WT}$ denote the coefficient selected
on the main WikiText validation set and frozen before this evaluation.
Each domain supplies 256 documents, with all next-token positions scored in
prefixes of at most 512 tokens. The decoder remains BF16; readouts use FP32.
No target-domain data are used to fit the quantizer or select $t_{\rm WT}$.
GPTQ uses a 65,536-state Hessian and the original 1,024-state range initializer.
Consequently, BLOOM's frozen GPTQ coefficient is $5$, whereas the original
1,024-state configuration in Table~\ref{tab:reparam-breadth} selects $4$.

Table~\ref{tab:cross-domain-transfer} reports all 24 comparisons. The frozen
coefficient improves over mean-centering in all 18 cases where $t_{\rm WT}\neq1$;
all corresponding paired 95\% intervals exclude zero. XGLM ties mean-centering
in six cases because $t_{\rm WT}=1$. All 24 frozen-coefficient KL point estimates
are below the unshifted baseline. Qwen3.5 illustrates that fixed mean-centering
can worsen fidelity: $t=1$ increases KL in every cell, while the frozen
$t_{\rm WT}=-0.5$ improves it.

\begin{table}[htbp]
\centering\scriptsize
\setlength{\tabcolsep}{2.5pt}
\caption{\textbf{Frozen WikiText coefficients beat centering in 18 transfer comparisons and tie in six.}
C4/OpenWebMath test KL to BF16 (nats): unshifted ($t=0$), centered ($t=1$),
and frozen ($t=t_{\rm WT}$). Frozen-minus-centered KL intervals use 10,000
paired document-bootstrap resamples; they are pointwise, conditional on
frozen coefficients, and unadjusted for multiple comparisons.}
\label{tab:cross-domain-transfer}
\resizebox{\textwidth}{!}{\input{generated/cross_domain_transfer.tex}}
\end{table}

\paragraph{Coefficient selection and validation size on BLOOM.}
A separate study evaluates all 14 coefficients on BLOOM, holding quantizer
fitting fixed. Each domain has 2,048 validation and 1,024 disjoint test documents,
with 32 sampled next-token positions per prefix. Table~\ref{tab:cross-domain-coefficients}
compares main WikiText coefficients with those selected on the full C4
and OpenWebMath validation pools. These coefficients are finite-grid selections
under unequal budgets, and should not be read as continuous or population optima.

\begin{table}[htbp]
\centering\small
\caption{\textbf{BLOOM's selected coefficient varies across validation domains.}
Column headings give validation sizes in articles/documents.
WikiText coefficients are from the main experiment; C4 and OpenWebMath selections use
separate 2,048-document pools.}
\label{tab:cross-domain-coefficients}
\input{generated/cross_domain_coefficients.tex}
\end{table}

We repeat selection using seven nested subset sizes ($16$--$1{,}024$ documents)
and ten seeds per domain, freezing each choice before test scoring.
Table~\ref{tab:cross-domain-stability} summarizes the smallest and largest sizes.
At 1,024 documents, all ten seeds agree for each of the six domain--quantizer
pairs, selecting $t=4$ or $t=5$. Several pairs already agree at 16 documents;
others vary among nearby grid candidates. The reported test KL averages the
ten selected heads' losses on the same fixed test set. This supports stable
non-unit choices for these two domains, without establishing universal stability:
in the broader five-language mC4 study, French RTN still selects $t=4$ in three
seeds and $t=5$ in seven at 1,024 documents.

\begin{table}[htbp]
\centering\scriptsize
\setlength{\tabcolsep}{3pt}
\caption{\textbf{At 1,024 validation documents, all ten seeds agree on a non-unit BLOOM coefficient per setting.}
Counts show selection frequencies. Test KL averages seeds on a fixed,
disjoint 1,024-document set unused for selection.}
\label{tab:cross-domain-stability}
\resizebox{\textwidth}{!}{\input{generated/cross_domain_stability.tex}}
\end{table}

\paragraph{Protocol distinctions.}
The transfer and stability studies use different document sets and position
sampling, so their absolute KL values should not be attributed solely to
coefficient selection. The transfer experiment freezes the selected coefficient
rather than a specific quantized tensor, and comparisons within each study use
matched reconstructed candidates. WikiText versus target-domain selection also
differs in validation-set size, so these experiments do not isolate distribution
shift from selection-budget and reconstruction effects.

%% file: generated/cross_domain_transfer.tex
\begin{tabular}{@{}lllrrrrr@{}}
\toprule
Head & Quantizer & Domain & $t_{\rm WT}$ & Raw KL & Centered KL & Frozen KL & $\Delta$ KL: 95\% CI \\
\midrule
Phi-4-mini & RTN & C4 & $4$ & 1.372879 & 0.899519 & 0.393513 & $[-0.517935,-0.493749]$ \\
Phi-4-mini & RTN & OpenWebMath & $4$ & 1.105418 & 0.675704 & 0.342138 & $[-0.348787,-0.318388]$ \\
Phi-4-mini & AW-MSE & C4 & $4$ & 1.039478 & 0.657195 & 0.304324 & $[-0.359930,-0.345035]$ \\
Phi-4-mini & AW-MSE & OpenWebMath & $4$ & 0.764193 & 0.498817 & 0.252586 & $[-0.256702,-0.235430]$ \\
Phi-4-mini & GPTQ & C4 & $4$ & 0.111289 & 0.078885 & 0.041984 & $[-0.037799,-0.035971]$ \\
Phi-4-mini & GPTQ & OpenWebMath & $4$ & 0.101269 & 0.072217 & 0.044054 & $[-0.029566,-0.026734]$ \\
BLOOM-1.7B & RTN & C4 & $6$ & 1.576803 & 1.226448 & 0.585626 & $[-0.669812,-0.611228]$ \\
BLOOM-1.7B & RTN & OpenWebMath & $6$ & 1.055612 & 0.872942 & 0.526070 & $[-0.371997,-0.322349]$ \\
BLOOM-1.7B & AW-MSE & C4 & $4$ & 0.943439 & 0.691491 & 0.147667 & $[-0.565035,-0.522129]$ \\
BLOOM-1.7B & AW-MSE & OpenWebMath & $4$ & 0.479431 & 0.396989 & 0.124050 & $[-0.293043,-0.253344]$ \\
BLOOM-1.7B & GPTQ & C4 & $5$ & 0.043011 & 0.044096 & 0.033579 & $[-0.011303,-0.009735]$ \\
BLOOM-1.7B & GPTQ & OpenWebMath & $5$ & 0.079911 & 0.071089 & 0.057823 & $[-0.014991,-0.011484]$ \\
XGLM-1.7B & RTN & C4 & $1$ & 1.833599 & 0.154811 & 0.154811 & $[+0.000000,+0.000000]$ \\
XGLM-1.7B & RTN & OpenWebMath & $1$ & 2.387851 & 0.159820 & 0.159820 & $[+0.000000,+0.000000]$ \\
XGLM-1.7B & AW-MSE & C4 & $1$ & 0.478122 & 0.098329 & 0.098329 & $[+0.000000,+0.000000]$ \\
XGLM-1.7B & AW-MSE & OpenWebMath & $1$ & 0.733782 & 0.095567 & 0.095567 & $[+0.000000,+0.000000]$ \\
XGLM-1.7B & GPTQ & C4 & $1$ & 0.009799 & 0.007307 & 0.007307 & $[+0.000000,+0.000000]$ \\
XGLM-1.7B & GPTQ & OpenWebMath & $1$ & 0.009308 & 0.006844 & 0.006844 & $[+0.000000,+0.000000]$ \\
Qwen3.5 & RTN & C4 & $-0.5$ & 0.029287 & 0.041919 & 0.022992 & $[-0.019664,-0.018170]$ \\
Qwen3.5 & RTN & OpenWebMath & $-0.5$ & 0.020447 & 0.029660 & 0.016817 & $[-0.013380,-0.012298]$ \\
Qwen3.5 & AW-MSE & C4 & $-0.5$ & 0.015426 & 0.025728 & 0.014890 & $[-0.011288,-0.010381]$ \\
Qwen3.5 & AW-MSE & OpenWebMath & $-0.5$ & 0.012674 & 0.019580 & 0.011050 & $[-0.008928,-0.008129]$ \\
Qwen3.5 & GPTQ & C4 & $-0.5$ & 0.008057 & 0.011161 & 0.007906 & $[-0.003389,-0.003122]$ \\
Qwen3.5 & GPTQ & OpenWebMath & $-0.5$ & 0.008031 & 0.011584 & 0.007309 & $[-0.004458,-0.004098]$ \\
\bottomrule
\end{tabular}

%% file: generated/cross_domain_coefficients.tex
\begin{tabular}{@{}lrrr@{}}
\toprule
Quantizer & WikiText (16) & C4 (2,048) & OpenWebMath (2,048) \\
\midrule
RTN & $6$ & $4$ & $4$ \\
AW-MSE & $4$ & $4$ & $5$ \\
GPTQ & $5$ & $4$ & $4$ \\
\bottomrule
\end{tabular}

%% file: generated/cross_domain_stability.tex
\begin{tabular}{@{}llllr@{}}
\toprule
Quantizer & Domain & Selection at $n=16$ & Selection at $n=1{,}024$ & Test KL: $16\to1{,}024$ \\
\midrule
RTN & C4 & $t=4$: 6/10; $t=6$: 4/10 & $t=4$: 10/10 & $0.572704\to 0.563011$ \\
RTN & OpenWebMath & $t=4$: 9/10; $t=6$: 1/10 & $t=4$: 10/10 & $0.532576\to 0.531784$ \\
AW-MSE & C4 & $t=4$: 10/10 & $t=4$: 10/10 & $0.146926\to 0.146926$ \\
AW-MSE & OpenWebMath & $t=4$: 3/10; $t=5$: 7/10 & $t=5$: 10/10 & $0.126947\to 0.126493$ \\
GPTQ & C4 & $t=3$: 3/10; $t=4$: 7/10 & $t=4$: 10/10 & $0.033628\to 0.033356$ \\
GPTQ & OpenWebMath & $t=3$: 5/10; $t=4$: 5/10 & $t=4$: 10/10 & $0.054443\to 0.053538$ \\
\bottomrule
\end{tabular}

%% file: generated/reparam_w23.tex
\begin{tabular}{llccc}
\toprule
Bits & Model & RTN & AW-MSE & GPTQ \\
\midrule
W2 & Phi & 98.2$\to$74.2 & 41.6$\to$16.2 & 3.08$\to$1.16 \\
W2 & Gemma 3 & 2.66$\to$2.21 & 0.822$\to$0.714 & 0.508$\to$0.491 \\
W2 & Gemma 4$^\dagger$ & 0.911$\to$0.911 & 0.212$\to$0.199 & 0.165$\to$0.162 \\
W2 & Qwen3.5 & 1.84$\to$1.74 & 0.312$\to$0.285 & 0.2$\to$0.197 \\
\cmidrule(lr){2-5}
W2 & BLOOM-1.7B & 16.7$\to$6.15 & 4.46$\to$1.46 & 0.427$\to$0.335 \\
W2 & BLOOMZ-1.7B & 16.2$\to$6.22 & 3.99$\to$2.12 & 0.456$\to$0.367 \\
W2 & XGLM-1.7B & 97.8$\to$55.3 & 23.7$\to$3.73 & 0.173$\to$0.128 \\
\midrule
W3 & Phi & 13$\to$2.46 & 4.62$\to$1.12 & 0.543$\to$0.211 \\
W3 & Gemma 3 & 0.244$\to$0.244 & 0.156$\to$0.144 & 0.112$\to$0.112 \\
W3 & Gemma 4$^\dagger$ & 0.0655$\to$0.062 & 0.0343$\to$0.0337 & 0.0352$\to$0.0341 \\
W3 & Qwen3.5 & 0.105$\to$0.0971 & 0.0566$\to$0.0523 & 0.0419$\to$0.0403 \\
\cmidrule(lr){2-5}
W3 & BLOOM-1.7B & 2.7$\to$1.72 & 1.02$\to$0.443 & 0.0969$\to$0.0839 \\
W3 & BLOOMZ-1.7B & 2.6$\to$1.58 & 0.943$\to$0.46 & 0.107$\to$0.0983 \\
W3 & XGLM-1.7B & 30$\to$0.926 & 14.2$\to$0.372 & 0.0352$\to$0.0265 \\
\bottomrule
\end{tabular}

%% file: generated/reparam_w23_ppl.tex
\begin{tabular}{lrccc}
\toprule
Head & BF16 & RTN & AW-MSE & GPTQ \\
\midrule
Phi & 9.73 & $3.98{\times}10^{6}$$\to$111.5 & 1002$\to$28.47 & 16.86$\to$11.97 \\
Gemma 3 & 38.49 & 44.34$\to$44.34 & 38.54$\to$38.89 & 40.09$\to$40.09 \\
Gemma 4$^\dagger$ & 66.41 & 71.11$\to$70.83 & 68.71$\to$69.06 & 69.05$\to$68.85 \\
Qwen3.5 & 8.98 & 9.851$\to$9.793 & 9.503$\to$9.446 & 9.351$\to$9.291 \\
\cmidrule(lr){1-5}
BLOOM-1.7B & 18.47 & 282.3$\to$100.7 & 50.78$\to$26.89 & 20.14$\to$20.04 \\
BLOOMZ-1.7B & 22.06 & 285.3$\to$124.9 & 56.08$\to$31.42 & 23.8$\to$23.9 \\
XGLM-1.7B & 13.80 & $1.53{\times}10^{14}$$\to$34.46 & $2.12{\times}10^{7}$$\to$20.27 & 14.19$\to$14.16 \\
\bottomrule
\end{tabular}

%% file: sections/app_stronger_controls.tex
\section{Complementarity and Stronger Controls on Phi}
\label{app:stronger-controls}

\paragraph{Protocol and scope.}
We recompute the controls on Phi-4-mini at W4/G128 with a frozen BF16 decoder,
128 fitting articles, 16 validation articles and 16 disjoint test articles,
using 512-token article prefixes and the same 14-point scalar grid as
Appendix~\ref{app:base-quantizers}. These are the main test articles,
not the untouched holdout of Appendix~\ref{app:holdout}. All choices are frozen
on validation KL before test evaluation. Source test PPL is 9.7192. The run
uses FP32 readouts with TF32 disabled, torch 2.12.0 and transformers 5.10.1;
its matched baselines are recomputed rather than pooled with earlier runs.
Table~\ref{tab:stronger-controls} gives the eight primary paired comparisons.
These comparisons are offline quality controls and do not measure packed serving or latency.

\paragraph{GPTQ calibration.}
The 1,024-state calibration set contains eight states from each fitting article.
The 65,536-state calibration set uses all 512 prefix positions from the same
articles to accumulate the full second moment. We retain 1\% damping, no
activation ordering and the pinned llmcompressor 0.12.0.1 solver. The middle
GPTQ condition changes only the Hessian while keeping the original AW-MSE
range statistics; the third also refits those statistics on the 65,536-state
calibration set.
All three conditions select $t=4$. Thus the gain survives both improved
error-feedback calibration and refitted initial ranges.

\paragraph{Exact channel scaling.}
For positive channel scales $s$, we quantize
$(W-t\mathbf1\mu^\top)\operatorname{diag}(s)$ and read out with
$\operatorname{diag}(s)^{-1}h$. Before quantization this preserves the source
softmax. The validation pool contains identity scaling, 20 activation-mean
power candidates $s_j\propto\E_{\rm fit}[|h_j|]^\alpha$ with
$\alpha\in\{0.05,0.10,\ldots,1\}$, and 11 SmoothQuant-style candidates
$s_j\propto\max_{\rm fit}|h_j|^\alpha/(\max_v|W_{vj}|)^{1-\alpha}$ with
$\alpha\in\{0,0.1,\ldots,1\}$. Scales are positive-clamped and normalized by
the geometric midpoint of their minimum and maximum; activation moments for
AW-MSE are transformed by $s_j^{-2}$. This adapts exact scaling to head-only
quantization and KL selection rather than reproducing a complete AWQ or
SmoothQuant pipeline. We first select scale-only, freeze that scale, and then
select the additive coefficient. Both quantizers select weight-max channel
equalization (the second family at $\alpha=0$); subsequent shifts select
$t=3$ for RTN and $t=4$ for AW-MSE. The search is sequential and does not sweep
the full scale--shift grid. Source equivalence is checked in FP64; we do not
claim numerically lossless folding into a BF16 normalization or measured
serving performance for the scaled variants.

\paragraph{Affine quantization and direction controls.}
Affine RTN uses group extrema including zero, a scale given by their range
divided by 15, a rounded integer zero point in $[0,15]$, and 16 reconstruction
levels. Symmetric RTN retains the paper's 15-level $[-7,7]$ grid with FP32
scales and reconstruction; AW-MSE and GPTQ use signed $[-8,7]$ codes, BF16
scales and BF16 reconstruction. Shift effects are compared within each
quantizer. The affine condition selects $t=5$. For direction controls, we
replace $\mu$ with the coordinate-wise median or one of three fixed-seed
Gaussian directions, each rescaled to $\|\mu\|_2$, and give every direction
the same scalar grid (Table~\ref{tab:stronger-directions}). Random shifts also
help, but the mean direction outperforms all tested alternatives; these
controls support its empirical utility rather than establish its optimality.

\paragraph{Uncertainty and evidence.}
Table~\ref{tab:stronger-controls-ci} reports paired article-bootstrap intervals
from 10,000 resamples (seed 20260912), recomputing token-weighted KL differences
within each sampled set of articles. The intervals are descriptive and
conditional on validation selection, with only 16 independent test articles.
They do not establish cross-family generalization. Experiment records retain
source revisions, token IDs, transform vectors, state/code hashes, frozen
selections, and per-article metrics.
\input{generated/reparam_stronger_controls.tex}
\clearpage

%% file: generated/reparam_stronger_controls.tex
\begin{table}[htbp]
\centering\small
\caption{\textbf{Phi shift gains persist under stronger calibration, scaling and affine quantization.} W4/G128 entries: baseline $\to$ shifted. Scaled baselines include selected channel scaling. BF16 PPL is 9.7192.}
\label{tab:stronger-controls}
\setlength{\tabcolsep}{4pt}
\begin{tabular}{lrrr}
\toprule
Condition & Test KL & Test PPL & KL reduction \\
\midrule
GPTQ: 1,024 states & $0.16041 \to 0.06051$ & $11.2883 \to 10.2858$ & $62.3\%$ \\
GPTQ: 65,536, original ranges & $0.08936 \to 0.03491$ & $10.5608 \to 10.0542$ & $60.9\%$ \\
GPTQ: 65,536, refitted ranges & $0.09247 \to 0.03491$ & $10.6275 \to 10.0182$ & $62.2\%$ \\
RTN & $1.23002 \to 0.35058$ & $34.1129 \to 13.4611$ & $71.5\%$ \\
AW-MSE & $0.93940 \to 0.25551$ & $24.8709 \to 12.4856$ & $72.8\%$ \\
Scaled RTN & $0.33659 \to 0.24234$ & $13.5410 \to 12.2376$ & $28.0\%$ \\
Scaled AW-MSE & $0.23998 \to 0.17199$ & $12.3055 \to 11.6372$ & $28.3\%$ \\
Affine RTN & $0.75002 \to 0.25878$ & $21.0794 \to 12.7901$ & $65.5\%$ \\
\bottomrule
\end{tabular}
\end{table}

\begin{table}[htbp]
\centering\small
\caption{\textbf{The mean direction gives the lowest KL among the tested alternatives on Phi.} Median and random directions are norm-matched to the mean and use the same validation scalar search. Random shifts also improve on the raw head.}
\label{tab:stronger-directions}
\setlength{\tabcolsep}{5pt}
\begin{tabular}{lrrrrrr}
\toprule
 & \multicolumn{3}{c}{RTN} & \multicolumn{3}{c}{AW-MSE} \\
\cmidrule(lr){2-4}\cmidrule(lr){5-7}
Direction & $t^\star$ & Test KL & Test PPL & $t^\star$ & Test KL & Test PPL \\
\midrule
Vocabulary mean & $4$ & 0.35058 & 13.4611 & $4$ & 0.25551 & 12.4856 \\
Coordinate-wise median & $4$ & 0.73863 & 20.1241 & $4$ & 0.55894 & 17.0431 \\
Random seed 0 & $-0.5$ & 1.05676 & 27.7650 & $-1$ & 0.77292 & 21.8672 \\
Random seed 1 & $2$ & 1.06612 & 28.1215 & $1$ & 0.86884 & 22.6155 \\
Random seed 2 & $-1$ & 1.04971 & 27.0636 & $-0.5$ & 0.84131 & 22.1993 \\
\bottomrule
\end{tabular}
\end{table}

\begin{table}[htbp]
\centering\small
\caption{\textbf{All eight Phi KL-reduction intervals exclude zero.} Negative differences favor shifting. Paired 95\% intervals use 10,000 bootstrap resamples of 16 test articles, conditional on validation selection.}
\label{tab:stronger-controls-ci}
\begin{tabular}{lrr}
\toprule
Condition & $\Delta$ test KL & $95\%$ interval \\
\midrule
GPTQ: 1,024 states & $-0.099903$ & $[-0.106154, -0.093710]$ \\
GPTQ: 65,536, original ranges & $-0.054452$ & $[-0.059731, -0.049574]$ \\
GPTQ: 65,536, refitted ranges & $-0.057559$ & $[-0.062531, -0.052901]$ \\
RTN & $-0.879442$ & $[-0.933053, -0.824182]$ \\
AW-MSE & $-0.683890$ & $[-0.726657, -0.637061]$ \\
Scaled RTN & $-0.094253$ & $[-0.104851, -0.084514]$ \\
Scaled AW-MSE & $-0.067993$ & $[-0.078665, -0.055833]$ \\
Affine RTN & $-0.491242$ & $[-0.521189, -0.464893]$ \\
\bottomrule
\end{tabular}
\end{table}

%% file: sections/app_meancentering.tex
\section{Mean-centering and representative selection}
\label{app:meancentering}

This section compares fixed mean-centering with validation selection and with
reconstruction-based selection. Data splits, candidate order and quantizer
conventions are defined in Appendix~\ref{app:base-quantizers}.

\subsection{Fixed centering versus the fidelity-selected shift}
\label{app:reparam-matrix}
Table~\ref{tab:reparam-matrix} supplies the selected coefficients and W4
RTN/AW-MSE values for the four main heads. The main-text
Table~\ref{tab:reparam-centering} compares these with $t=1$ for the fragile heads.

\begin{table}[ht]
\centering\small
\caption{\textbf{Phi benefits most from W4 shifts; the other three heads change little.}
RTN and AW-MSE use G128 and independently select coefficients by validation KL.
Entries report raw and reparameterized test KL against each head's source
distribution. Gemma~4 ($\dagger$) uses the rank-one soft-cap correction.}
\label{tab:reparam-matrix}
\input{generated/reparam_matrix.tex}
\end{table}

Transferring the RTN-selected coefficient to AW-MSE raises Gemma~3 test KL to
0.042, compared with 0.038 under AW-MSE's own selection and 0.041 without a shift.

\subsection{Reconstruction-selected versus fidelity-selected representative}
\label{app:reparam-scan}
The exploratory reconstruction scan uses symmetric RTN at W2, W3 and W4
and $t\in\{0,0.25,0.5,0.75,1,1.25,1.5,2\}$. For
$E_t=Q_t-W_t$ and its vocabulary-row mean
$\bar e_t=E_t^\top\mathbf1/V$, it measures squared weight error and its
projection onto softmax-visible row differences:
\begin{equation}
 R_{\rm raw}(t)=\|E_t\|_F^2,\qquad
 R_{\rm proj}(t)=\|PE_t\|_F^2
 =R_{\rm raw}(t)-V\|\bar e_t\|_2^2.
 \label{eq:reparam-projected-weight}
\end{equation}
The implementation divides both quantities by $Vd$, which leaves their
rankings unchanged. This quantity is a projected \emph{weight} MSE; it does not
incorporate the logit reconstruction $\E\|PE_t h\|^2$ or a Fisher-weighted objective.
The removed term measures a shared residual component; it does not supply
hidden-state covariance or output-probability weighting.

The retained sweep records cover three models, three precisions and eight
coefficients per comparison. Both reconstruction criteria select $t=1$ in
all nine comparisons, while the coefficient minimizing diagnostic KL differs
in every case (Table~\ref{tab:reparam-scan}). At W4, KL favors $t=0.5$ for
Gemma, $t=0$ for Qwen and $t=2$ for Phi within this grid. Projecting away the
common residual component therefore does not recover fidelity-based selection.

\begin{table}[ht]
\centering\small
\caption{\textbf{Weight MSE and KL favor different shifts.}
Each row uses the same symmetric RTN quantizer and diagnostic states at every
coefficient. Both raw and projected weight MSE select $t=1$; the KL minimum
is descriptive on this diagnostic set, not independently selected on validation.}
\label{tab:reparam-scan}
\input{generated/reparam_scan.tex}
\end{table}

The two weight-space criteria fail on these tested candidates as measured;
this does not extend to all reconstruction objectives. The scan
stops at $t=2$ and cannot locate the later $t=4$ choice. The wider experiment
supplies the separate validation-selected comparison. All rows use 32 test
article prefixes and FP32 linear-logit evaluation; per-candidate aggregates
are retained, while source and captured-state hashes require additional
provenance. Their RTN convention differs from the signed-scale centering
control below, so the two tables are not evaluations of identical artifacts.

\subsection{Range and shared-row controls}
\label{app:reparam-centering-controls}
A separate signed min--max control compares $t=0$ and $t=1$ with BF16
stored scales and reconstructed weights, followed by FP32 logit evaluation.
Table~\ref{tab:mean-centering-diag} reports its W4 results. The shared weight
energy is $V\|\mu\|^2/\|W\|_F^2$, and the range ratio compares the mean
group maximum absolute weight after and before centering. Smaller ranges
alone do not explain the outcome: Gemma and Qwen have smaller mean ranges
yet worse KL, whereas Phi has a slightly larger mean range and better KL.
Qwen and Phi also have comparable shared weight-energy fractions but
opposite fidelity changes.

\begin{table}[ht]
\centering\small
\caption{\textbf{Smaller group ranges do not guarantee lower KL.}
Shared energy and range ratios describe unquantized weights.
KL compares raw and mean-centered W4 G128 signed min--max heads on matched
diagnostic articles.}
\label{tab:mean-centering-diag}
\input{generated/reparam_centering.tex}
\end{table}

%% file: generated/reparam_matrix.tex
\begin{tabular}{lrrrrrr}
\toprule
 & \multicolumn{3}{c}{RTN} & \multicolumn{3}{c}{AW-MSE} \\
Model & $t^\star$ & Raw KL & Reparam KL & $t^\star$ & Raw KL & Reparam KL \\
\midrule
Phi & 4 & 1.229 & 0.351 & 4 & 0.936 & 0.256 \\
Gemma 3 & 0.5 & 0.049 & 0.046 & $-1$ & 0.041 & 0.038 \\
Gemma 4$^\dagger$ & 1 & 0.012 & 0.010 & 0.5 & 0.008 & 0.008 \\
Qwen3.5 & $-0.5$ & 0.021 & 0.019 & $-0.5$ & 0.013 & 0.013 \\
\bottomrule
\end{tabular}

%% file: generated/reparam_scan.tex
\begin{tabular}{lrrrrrr}
\toprule
 & & \multicolumn{3}{c}{Best sampled $t$} & \multicolumn{2}{c}{Diagnostic KL} \\
Model & Bits & MSE & Proj. MSE & KL & At MSE min & At KL min \\
\midrule
Gemma 3 & 4 & 1 & 1 & 0.5 & 0.0580 & 0.0465 \\
Gemma 3 & 3 & 1 & 1 & 0.25 & 0.3161 & 0.2340 \\
Gemma 3 & 2 & 1 & 1 & 0 & 2.8305 & 2.5793 \\
Qwen3.5 & 4 & 1 & 1 & 0 & 0.0291 & 0.0202 \\
Qwen3.5 & 3 & 1 & 1 & 0 & 0.1599 & 0.1017 \\
Qwen3.5 & 2 & 1 & 1 & 0 & 2.6671 & 1.8072 \\
Phi & 4 & 1 & 1 & 2 & 0.7981 & 0.5644 \\
Phi & 3 & 1 & 1 & 2 & 8.9786 & 5.3672 \\
Phi & 2 & 1 & 1 & 1.5 & 77.0395 & 74.4024 \\
\bottomrule
\end{tabular}

%% file: generated/reparam_centering.tex
\begin{tabular}{lrrrr}
\toprule
Model & Shared weight energy (\%) & Range ratio & Raw KL & Centered KL \\
\midrule
Gemma 3 & 3.41 & 0.9612 & 0.0499 & 0.0585 \\
Qwen3.5 & 14.23 & 0.9029 & 0.0196 & 0.0294 \\
Phi & 16.40 & 1.0128 & 1.2109 & 0.8087 \\
\bottomrule
\end{tabular}

%% file: sections/app_mechanism.tex
\section{Mechanism Details}
\label{app:mechanism}

On the matched Phi diagnostics, the selected shift increases reconstruction
error while reducing distributional error (Section~\ref{sec:reparam-analysis}).
This appendix defines the probability-bin attribution, tests the residual
pattern under stronger GPTQ calibration, and gives a separate cross-model
comparison of output sensitivity.

\subsection{Fisher and probability-bin diagnostics}
\label{app:reparam-matched-mechanism}
The paired diagnostic uses the source-matched Phi state set, its 1,024 fitting
states, and the last 16 of its 32 held-out articles (8,176 prediction states).
The coefficients $t\in\{0,1,4\}$ are fixed; no selection occurs on these states.
For each signed W4 G128 RTN/AW-MSE head, the probe retains the quantized-weight
hash and computes actual KL and the Fisher quadratic from the same residual
$e=E_t h$, with FP32 products and TF32 disabled. AW-MSE uses the twelve-factor
clipping grid of the probability-probe protocol. The source, hidden-state,
fitting-state, artifact and code hashes accompany the measurements.

Both quantizers use signed integers, BF16 stored scales and BF16
reconstructed weights. The RTN convention differs from the symmetric
FP32-scale selection probe, so all metrics are recomputed for this comparison.

To understand where the additional reconstruction error goes, we group
vocabulary entries by their source probabilities. We then measure how much
reconstruction error and Fisher-weighted error each group contributes, using
the same groups for every shift.

For each source distribution $p$, we attribute squared reconstruction energy using
$e_c=e-\operatorname{mean}_v(e)\mathbf1$ and Fisher energy using
$p_v(e_v-p^\top e)^2$. Summing bins recovers $\|Pe\|^2$ and $e^\top F(p)e$,
respectively. Error energy in Table~\ref{tab:reparam-probability-bins} is the
mean per-state sum within a bin; shares divide aggregate bin energy by
aggregate projected energy. Probability mass is averaged across states.
The lowest-probability bin contains 98.75\% of vocabulary entries on average,
so its large error share alone is not evidence of preferential concentration.
The evidence is the change in shares across coefficients on the same states,
together with the decline in absolute error on high-probability entries.
Vocabulary centering removes 4.1\% of AW-MSE residual energy at $t=0$ and
27.5\% at $t=4$; the RTN shares are 0.03\% and 5.7\%.

\begin{table}[ht]
\centering\scriptsize
\setlength{\tabcolsep}{3pt}
\caption{\textbf{Phi's selected shift reduces error on high-probability outputs.}
Bins use unchanged source probabilities on matched states.
Error is vocabulary-centered logit-error energy; Fisher shares use
probability-centered residuals.}
\label{tab:reparam-probability-bins}
\input{generated/reparam_probability_bins.tex}
\end{table}

\subsection{Residual diagnostics under stronger GPTQ}
\label{app:gptq-mechanism}
We repeat the residual diagnostics for Phi W4 G128 GPTQ using the full
Hessian calibrated on 65,536 states, while retaining the original AW-MSE
range initializer fitted on 1,024 states. This is the original-range
condition in Table~\ref{tab:stronger-controls}, with 1\% damping and no
activation ordering. We compare $t=0$ with its previously frozen,
validation-selected $t^\star=4$; no further coefficient search is performed.

The two rows use the same 8,176 held-out states from 16 articles, with FP32
products and log-softmax, TF32 disabled, and BF16 scales and reconstructed
weights. These states exactly reproduce the stronger-GPTQ evaluation.
The GPTQ and RTN/AW-MSE diagnostics use separately captured hidden-state
tensors from the same source weights and article identities; all comparisons
are paired within their respective capture.

\begin{table}[htbp]
\centering\small
\setlength{\tabcolsep}{5pt}
\caption{\textbf{Shifting Phi W4 GPTQ lowers KL despite greater logit-error energy.}
GPTQ uses a 65,536-state full Hessian and the original range initializer.
Definitions follow Table~\ref{tab:reparam-mechanism}; test states are matched.}
\label{tab:gptq-mechanism}
\input{generated/phi_gptq_mechanism.tex}
\end{table}

From $t=0$ to $t=4$, relative weight error rises by 9.2\%, raw logit-error
energy rises by $2.66\times$, and vocabulary-centered error energy rises by
$2.55\times$. Nevertheless, Fisher-weighted error falls by 60.90\% and KL
by 60.94\%. The quadratic is within 0.7\% of actual KL in both rows.
Paired article-bootstrap 95\% intervals for selected minus raw are
$[-0.05975,-0.04954]$ for KL and $[-0.05982,-0.04884]$ for Fisher/2
(10,000 resamples, conditional on the frozen selection).
Thus the qualitative residual pattern persists after stronger GPTQ error
compensation on this Phi W4 comparison. We do not infer a universal mechanism
across models or extend the local quadratic claim to W2.

\subsection{Cross-model output sensitivity}
\label{sec:output-geometry}
Weight-space error is not a reliable proxy for prediction fidelity. On a matched
Gemma W3 G128 control, activation weighting \emph{raises} relative weight $L_2$
error from 0.2063 to 0.2162 while \emph{reducing} held-out KL from 0.24443 to
0.14194: weight-space distortion becomes logit error only through the occupied
hidden-state directions, and the prediction distribution reweights that error.
This motivates selecting representatives by prediction fidelity rather than by
weight norm.

For the cross-model comparison, let $\delta z=(Q-W)h$ and use the categorical
Fisher matrix and local KL expansion from Equation~\ref{eq:reparam-fisher}.
For nonzero aggregate error, the per-unit sensitivity is
$C_{\rm out}=\E[\delta z^\top F(p)\delta z]/\E\|\delta z\|_2^2$, and
$S_p=2\E[D_{\rm KL}]/\E\|\delta z\|_2^2$ is the actual KL per unit logit error.
Table~\ref{tab:output-geometry} compares W4 AW-MSE heads fitted to the same
articles across Gemma, Qwen and Phi, with source-normalized distortion
$D_z=\E\|\delta z\|^2/\E\|z\|^2$.
Phi has smaller source-normalized distortion $D_z$ but roughly $12\times$
larger absolute logit-error energy and about $6\times$ higher per-unit Fisher
sensitivity than Qwen; both factors contribute to its much larger KL.
Phi also has a smaller mean
top-two logit margin than Gemma (1.96 versus 2.59) and a larger mean Fisher
trace (0.592 versus 0.452). These measurements describe the evaluated unshifted heads; they do not
establish a general predictor of the benefit from reparameterization.

\begin{table}[t]
\centering\small
\caption{\textbf{Phi exceeds Qwen in both logit-error energy and sensitivity.}
Unshifted W4 AW-MSE uses disjoint WikiText articles (16,352 positions/model).
$D_z$ normalizes error by source energy; $S_p$ measures sensitivity per unit
error and $C_{\rm out}$ its Fisher approximation on the same residuals.}
\label{tab:output-geometry}
\input{generated/output_geometry.tex}
\end{table}

%% file: generated/reparam_probability_bins.tex
\begin{tabular}{lllrrrr}
\toprule
Quantizer & $t$ & Source probability & Mass (\%) & Error share (\%) & Error energy & Fisher share (\%) \\
\midrule
RTN & 0 & $p<10^{-6}$ & 0.196 & 93.474 & 134039.57 & 0.317 \\
RTN & 0 & $10^{-6}\leq p<10^{-4}$ & 2.196 & 6.024 & 8637.93 & 4.015 \\
RTN & 0 & $10^{-4}\leq p<10^{-2}$ & 12.842 & 0.476 & 682.63 & 24.223 \\
RTN & 0 & $p\geq10^{-2}$ & 84.766 & 0.026 & 36.88 & 71.444 \\
RTN & 1 & $p<10^{-6}$ & 0.196 & 94.179 & 106621.80 & 0.392 \\
RTN & 1 & $10^{-6}\leq p<10^{-4}$ & 2.196 & 5.381 & 6092.47 & 4.994 \\
RTN & 1 & $10^{-4}\leq p<10^{-2}$ & 12.842 & 0.421 & 476.38 & 28.570 \\
RTN & 1 & $p\geq10^{-2}$ & 84.766 & 0.019 & 21.78 & 66.044 \\
RTN & 4 & $p<10^{-6}$ & 0.196 & 99.074 & 395859.36 & 0.445 \\
RTN & 4 & $10^{-6}\leq p<10^{-4}$ & 2.196 & 0.862 & 3444.53 & 4.941 \\
RTN & 4 & $10^{-4}\leq p<10^{-2}$ & 12.842 & 0.061 & 243.17 & 27.934 \\
RTN & 4 & $p\geq10^{-2}$ & 84.766 & 0.003 & 10.37 & 66.680 \\
AW-MSE & 0 & $p<10^{-6}$ & 0.196 & 92.436 & 132569.58 & 0.348 \\
AW-MSE & 0 & $10^{-6}\leq p<10^{-4}$ & 2.196 & 7.004 & 10044.88 & 4.162 \\
AW-MSE & 0 & $10^{-4}\leq p<10^{-2}$ & 12.842 & 0.535 & 767.86 & 25.173 \\
AW-MSE & 0 & $p\geq10^{-2}$ & 84.766 & 0.024 & 35.00 & 70.316 \\
AW-MSE & 1 & $p<10^{-6}$ & 0.196 & 92.686 & 106162.10 & 0.403 \\
AW-MSE & 1 & $10^{-6}\leq p<10^{-4}$ & 2.196 & 6.773 & 7757.36 & 4.781 \\
AW-MSE & 1 & $10^{-4}\leq p<10^{-2}$ & 12.842 & 0.518 & 592.84 & 27.345 \\
AW-MSE & 1 & $p\geq10^{-2}$ & 84.766 & 0.023 & 26.70 & 67.471 \\
AW-MSE & 4 & $p<10^{-6}$ & 0.196 & 98.201 & 288852.93 & 0.484 \\
AW-MSE & 4 & $10^{-6}\leq p<10^{-4}$ & 2.196 & 1.679 & 4938.41 & 5.228 \\
AW-MSE & 4 & $10^{-4}\leq p<10^{-2}$ & 12.842 & 0.116 & 341.06 & 27.828 \\
AW-MSE & 4 & $p\geq10^{-2}$ & 84.766 & 0.005 & 13.48 & 66.460 \\
\bottomrule
\end{tabular}

%% file: generated/phi_gptq_mechanism.tex
\begin{tabular}{rrrrrr}
\toprule
$t$ & $\|E_t\|/\|W_t\|$ & $D_z$ (\%) & Common (\%) & Fisher/2 & Actual KL \\
\midrule
0 & 0.148515 & 0.025340 & 0.4644 & 0.088771 & 0.089362 \\
4 & 0.162178 & 0.067442 & 4.7268 & 0.034710 & 0.034909 \\
\bottomrule
\end{tabular}

%% file: generated/output_geometry.tex
\begin{tabular}{lrrrrr}
\toprule
Model & $\E\|\delta z\|^2$ & $D_z$ & $S_p$ & $C_{\mathrm{out}}$ & KL \\
\midrule
Gemma 3 4B & $2.08\times10^{4}$ & $2.93\times10^{-3}$ & $3.89\times10^{-6}$ & $3.94\times10^{-6}$ & 0.041 \\
Qwen3.5 4B & $1.23\times10^{4}$ & $7.44\times10^{-3}$ & $2.09\times10^{-6}$ & $2.09\times10^{-6}$ & 0.013 \\
Phi-4-mini & $1.47\times10^{5}$ & $2.41\times10^{-3}$ & $1.26\times10^{-5}$ & $1.19\times10^{-5}$ & 0.925 \\
\bottomrule
\end{tabular}

%% file: sections/app_deployment.tex
\section{Deployment and Systems Validation}
\label{app:deployment-systems}

This section gives the matched packed Phi export and timing protocol behind the
main deployment table, the tied/untied storage accounting, and an auxiliary Qwen
serving measurement. Packed checkpoints use the original parameterization unless
the reparameterized export is stated; each numerical path retains its own BF16
reference.

\subsection{Matched Phi packed export and timing}
\label{app:reparam-integrated}
The deployment retest starts from one BF16 Phi source and independently
builds untied BF16 heads at $t=0,4$ and W4 G128 heads at $t=0,4$ for min--max
and AW-MSE. The exporter resolves the source output matrix, preserves all 194
non-head tensors byte-for-byte, and verifies every serialized head tensor.
For the tied source, the BF16 input embedding is retained. Every W4 log records
selection of the Marlin linear kernel in vLLM 0.28.0 on an A10G.

Quality uses 32 contiguous 512-token WikiText blocks, scoring 16,352 next
tokens with eager execution and a 1,024-token model limit. Corpus and token
hashes match across all seven rows. Per-block sums reproduce NLL and its
exponential reproduces PPL. These fixed deployment evaluation blocks are distinct
from the article-partitioned selection/evaluation split and were not held out as
a fresh test set. Table~\ref{tab:reparam-integrated} isolates the BF16 serialization controls.

\begin{table}[ht]
\centering\small
\caption{\textbf{Untying and BF16 serialization leave small likelihood differences on Phi.}
Matched exports test the unshifted and shifted heads; algebraic softmax
invariance does not imply bitwise inference equivalence.}
\label{tab:reparam-integrated}
\input{generated/reparam_deployment_controls.tex}
\end{table}

The selected $t=4$ is fixed before deployment evaluation for both base
quantizers. AW-MSE uses second moments from the 1,024 fitting states used for this
deployment configuration, whose source checkpoint identity matches the
deployment reference. The
exporter computes the row mean and shift in FP32, casts the representative
to BF16, then selects signed-integer codes using floating-point candidate
scales and stores BF16 scales. Its clipping grid is
$\{1,.975,.95,.925,.9,.875,.85,.8,.75,.7\}$.
This deployment path differs from the
probability probes' scale-rounding and fitting conventions. No coefficient
or clipping grid is selected using the reported deployment PPL.

Timing uses greedy 64-token generation at batches one and 16, with prefix
caching disabled and normal graph execution. Each cell has three warmups
and seven timed repetitions. We independently reload the five source/W4
checkpoints in two rounds, reversing their order in the second round.
The main table reports the mean of the two round medians; timing includes
prefill and generation and excludes loading. GPU access is exclusive during
the run. Checkpoint hashes match between every quality and timing row.
Shifted AW-MSE batch-one medians are 1,013.24 and 1,013.53\,ms; its B16
medians are 1,210.19 and 1,210.39\,ms. These repeated measurements show
essentially unchanged W4 latency; they do not guarantee zero overhead on other
workloads or devices. The matched min--max exports reproduce the 40.57 to 16.28 PPL reduction.
A deployed rank-one correction for Gemma~4 is not part of these measurements.

\paragraph{GPTQ construction and timing scope.}
The deployed full-Hessian GPTQ head uses the same INT4 Marlin format as AW-MSE.
Its construction takes about one minute per shifted head, compared with a few
seconds for AW-MSE scale selection; Appendix~\ref{app:selection-protocol} reports
full sweep costs. GPTQ PPL and KL are evaluated separately. GPTQ was not timed,
so its latency entries in Table~\ref{tab:reparam-deployment} are left unreported.

\subsection{Tied and untied storage}
\label{app:deployment}
\paragraph{Head payload.}
With $Vd$ indices and one BF16 scale per $G$ weights, the dominant payload is
\begin{equation}
 M_{\rm head}(b,G)=Vd\left(\frac b8+\frac2G\right)\ {\rm bytes}.
 \label{eq:payload}
\end{equation}
This excludes small metadata and does not account for any retained embedding.

\paragraph{Model interface.}
A model adapter identifies the output matrix, final normalized hidden
states, embedding-sharing configuration, and any transformation between
linear logits and probabilities. For an already untied head, the packed output
matrix replaces the existing head. For a tied head, the evaluated layout retains
the original BF16 embedding and adds an independently packed output projection.
A runtime that supports a shared quantized embedding and projection would require
separate evaluation of both operators and their combined quality impact.
The current head-only comparisons do not evaluate that alternative.

\paragraph{Storage and memory traffic.}
Table~\ref{tab:payload} instantiates Equation~\ref{eq:payload} for the
evaluated checkpoint. In an untied model, replacing an independent BF16 head
reduces its weight storage by $2Vd-M_{\rm head}$. In the evaluated tied
layout, preserving the BF16 embedding instead adds $M_{\rm head}$ of
resident weights. The tied Phi source loads about 7.17\,GiB of model weights,
versus 7.47\,GiB with the independent packed head: lower output-projection
traffic coexists with higher resident weight memory. Actual memory
traffic depends on caching, batch reuse, packing and kernel execution;
latency also includes the decoder, output processing and sampling.
The payload calculation does not measure runtime workspaces or the KV cache.

\begin{table}[H]
\centering\small
\caption{\textbf{W4 G128 uses 26\% of the BF16 head payload.}
Analytical Gemma~3 payload: $V=262{,}208$, $d=2{,}560$.
An untied packed head adds resident storage when the BF16 input embedding
is retained.}
\label{tab:payload}
\begin{tabular}{lrr}
\toprule
Representation & Head payload (GiB) & Relative to BF16 \\
\midrule
BF16 & 1.2503 & 1.0000 \\
W8, G128 & 0.6349 & 0.5078 \\
W4, G128 & 0.3223 & 0.2578 \\
W4, G32 & 0.3516 & 0.2812 \\
\bottomrule
\end{tabular}
\end{table}

\paragraph{Packing and loading.}
For the packed W4/W8 configurations, the checkpoint builder stores signed
quantized codes in INT32 containers with $32/b$ codes per container, together
with BF16 group scales and the original matrix shape. It exports an independent
compressed-tensors head group and uses the target pattern
\texttt{re:.*lm\_head\$} to match the mapped output operator. Tied-model export
clears the applicable embedding-sharing flags while retaining the source input
table. A deployment audit should bind the fitted artifact, serialized tensors and
loaded operator to the same weights and stored scales, verify the signed code
convention through a packing round trip, and record actual kernel dispatch. The
intended serving path uses vLLM and a compatible packed projection such as Marlin
\citep{kwon2023vllm,marlinartifact}; quantization metadata or reduced file size
alone does not establish that this path ran.

\subsection{Auxiliary Qwen serving}
\label{app:cross-model}
The Qwen3.5 4B and Phi-4-mini checkpoints used above quantize BF16-decoder heads,
retain BF16 input embeddings and export independent W4/W8 G128 heads. Fitting uses
128 generic prompts with eight sampled positions each. The Qwen and Phi KL and
top-1 diagnostics use the 1,024 fitting states: at W4, AW-MSE improves Qwen top-1
agreement from 93.26\% to 95.21\% and Phi from 37.01\% to 49.12\% (in-distribution
diagnostics on the fitting states, without exhaustive non-head tensor audits).

\begin{table}[ht]
\centering\small
\caption{\textbf{W4 reduces Qwen3.5 4B batch-one latency by 9.6\% versus BF16 on A10G.}
Head-only W4/W8 use G128; timings cover 25-token greedy generation over seven
warmed compiled-vLLM repetitions, excluding loading and compilation.}
\label{tab:qwen-serving}
\input{generated/qwen_serving.tex}
\end{table}

Table~\ref{tab:qwen-serving} reports medians from persistent engines. The kernel
audit records vLLM 0.28.0 with compiled CUDA graphs and
\texttt{MarlinLinearKernel} for Qwen's packed heads. Generation timing includes
the output projection and surrounding model execution, rather than an isolated
head microbenchmark; the 9.6\% batch-one latency reduction is $1-467.2/516.6$
using unrounded measurements. Warm persistent engines exclude loading and
compilation, and the interval does not separately identify prefill and decode
costs or variation across independent engine starts.

%% file: generated/reparam_deployment_controls.tex
\begin{tabular}{lrr}
\toprule
BF16 head & WikiText PPL & $\Delta$NLL vs. source \\
\midrule
Tied source & 11.6477 & +0.000000 \\
Untied, $t=0$ & 11.6455 & -0.000194 \\
Untied, $t=4$ & 11.6363 & -0.000983 \\
\bottomrule
\end{tabular}

%% file: generated/qwen_serving.tex
\begin{tabular}{lrr}
\toprule
Head & B1 latency (ms) & B16 throughput (tokens/s) \\
\midrule
BF16 & 516.6 & 570.9 \\
W8 AW-MSE & 480.4 & 585.9 \\
W4 AW-MSE & 467.2 & 614.8 \\
\bottomrule
\end{tabular}

%% file: sections/app_figdetails.tex
\section{Model, Head and Vocabulary Details}
\label{app:checkpoints}

These tables support Figure~\ref{fig:head-size} and the cross-model precision
comparison; they are reference material rather than part of the method.

\subsection{Output-head sizes}
\label{app:architectures}
Table~\ref{tab:architectures} gives the dimensions of the evaluated heads. Head parameters are $Vd$; the nominal share
divides this count by the model size in the name (3.8B for Phi-4-mini).
It indicates scale and does not represent a fraction of resident memory. Tied embeddings
and multimodal components require separate accounting. Gemma~4 E4B is
included for its output dimensions, without assigning a nominal share:
its effective-size designation is not a total parameter count.
The Gemma~3 row uses the evaluated checkpoint's vocabulary dimension;
the other rows use official configurations
\citep{gemma4config,qwen35config,phi4miniconfig}.
The three additional very-large-vocabulary heads (BLOOM, BLOOMZ, XGLM) are 1.7B
models whose 250--256K vocabularies place roughly $30\%$ of nominal parameters in
the output head.

\begin{table}[ht]
\centering\small
\caption{Output-projection dimensions and nominal parameter shares.
Gemma~4's 262,144-row configuration and the evaluated Gemma~3 checkpoint's
262,208-row matrix both round to 671M head weights.}
\label{tab:architectures}
\input{generated/head_size_table.tex}
\end{table}

\paragraph{Vocabulary comparison and sources.}
Figure~\ref{fig:head-size} compares 23 selected model families using output-matrix
row counts, including reserved or padded IDs. Earlier bars are shown for 16
families; seven have only a selected checkpoint. The paired examples do not
establish monotonic growth across intermediate releases. The GPT row groups
a vendor lineage across architectural changes. The right panel uses each selected checkpoint's own hidden
width and decimal GB; its fourfold reduction describes raw weight codes only.
Table~\ref{tab:vocab-checkpoints} identifies all 23 selected checkpoints and the
16 earlier references, with linked configurations and release sources. The
Gemma~3 row uses the evaluated 262,208-row matrix from
Table~\ref{tab:architectures}; native Gemma configurations can use 262,144.
Machine-readable plotted values and the full source record accompany the paper.
Table~\ref{tab:fig1a-vocab} provides additional intermediate generations for four
families. Each is read from the
model's official configuration (the \texttt{vocab\_size} field of
\texttt{config.json} on the Hugging Face Hub); the corresponding family reports
and cards include
\citep{llama3report,qwen25report,qwen35config,phi4miniconfig,ministral3instructconfig}.

\input{figdata/vocabulary_checkpoints.tex}

\begin{table}[ht]
\centering\small
\caption{Vocabulary sizes across selected generations of four families, from each
model's official configuration.}
\label{tab:fig1a-vocab}
\begin{tabular}{llr}
\toprule
Family & Model & Vocabulary \\
\midrule
Phi & Phi-2 & 51{,}200 \\
 & Phi-3-mini & 32{,}064 \\
 & Phi-4 & 100{,}352 \\
 & Phi-4-mini & 200{,}064 \\
\midrule
Llama & Llama~2 & 32{,}000 \\
 & Llama~3 & 128{,}256 \\
 & Llama~4 & 202{,}048 \\
\midrule
Qwen & Qwen1--2.5 & 151{,}936 \\
 & Qwen3.5 & 248{,}320 \\
\midrule
Mistral & Mistral~7B & 32{,}000 \\
 & Ministral~3 & 131{,}072 \\
\bottomrule
\end{tabular}
\end{table}

\subsection{Precision and perplexity across models}
\label{sec:cross-model}
Similarly sized heads have different precision requirements
(Table~\ref{tab:cross-model}). W4 AW-MSE keeps perplexity within 1.1\% of
each model's BF16 reference for Gemma 3/4 and Qwen3.5. Phi is
the exception: weighting reduces its W4 PPL from 40.57 to 30.94, still far
above BF16's 11.65; W8 AW-MSE reaches 11.75. Among the evaluated settings,
W8 AW-MSE preserves near-reference likelihood for Phi, while W4 AW-MSE
suffices for the other three models. The Qwen--Phi contrast holds group size fixed and establishes
that matrix size alone does not determine the observed precision requirement;
distributional diagnostics on their fitting states show the same contrast
(W4 AW-MSE KL 0.01361 for Qwen versus 0.88979 for Phi), examined further in
Appendix~\ref{sec:output-geometry}. These observations compare practical
operating points rather than locating each model's lowest viable precision.

\begin{table}[H]
\centering\small
\caption{\textbf{W4 AW-MSE retains near-BF16 perplexity for three heads but substantially degrades Phi.}
Head-only WikiText evaluation uses G32 for Gemma~3 W4 and G128 otherwise.
Each model uses its own BF16 reference and tokenizer, with matched Hugging Face
loading and evaluation paths.}
\label{tab:cross-model}
\input{generated/cross_model_likelihood.tex}
\end{table}

%% file: generated/head_size_table.tex
\begin{tabular}{lrrrr}
\toprule
Model & Vocabulary & Width & Head (M) & Nominal share (\%) \\
\midrule
Gemma 3 4B & 262,208 & 2,560 & 671 & 16.8 \\
Gemma 4 E4B & 262,144 & 2,560 & 671 & --- \\
Qwen3.5 4B & 248,320 & 2,560 & 636 & 15.9 \\
Phi-4-mini & 200,064 & 3,072 & 615 & 16.2 \\
\midrule
BLOOM-1.7B & 250,880 & 2,048 & 514 & 30.2 \\
BLOOMZ-1.7B & 250,880 & 2,048 & 514 & 30.2 \\
XGLM-1.7B & 256,008 & 2,048 & 524 & 30.8 \\
\bottomrule
\end{tabular}

%% file: figdata/vocabulary_checkpoints.tex
\begingroup
\hypersetup{hidelinks}
\scriptsize
\def\UrlBreaks{\do\/\do-\do\_\do.}
\setlength{\tabcolsep}{3pt}
\renewcommand{\arraystretch}{1.08}
\begin{longtable}{@{}>{\raggedright\arraybackslash}p{.52in}>{\raggedright\arraybackslash}p{1.94in}>{\raggedright\arraybackslash}p{1.94in}r@{}}
\caption{Checkpoints behind Figure~\ref{fig:head-size}, in plotted order. Checkpoint names link to configuration sources and years to release sources. $V$ counts output rows; $d$ is the selected checkpoint width. A dash indicates no earlier comparison.}\label{tab:vocab-checkpoints}\\
\toprule
Family & Earlier checkpoint and rows & Selected checkpoint and rows & $d$ \\
\midrule
\endfirsthead
\multicolumn{4}{l}{\tablename~\thetable\ (continued)}\\
\toprule
Family & Earlier checkpoint and rows & Selected checkpoint and rows & $d$ \\
\midrule
\endhead
\midrule
\multicolumn{4}{r}{Continued on next page}\\
\endfoot
\bottomrule
\endlastfoot
Gemma & \href{https://github.com/google/gemma_pytorch/blob/main/gemma/config.py}{\nolinkurl{google/gemma-2b}}\newline [\href{https://blog.google/innovation-and-ai/technology/developers-tools/gemma-open-models/}{2024}]; $V=256{,}000$ & \hyperref[tab:architectures]{\nolinkurl{google/gemma-3-4b-it}}\newline [\href{https://blog.google/innovation-and-ai/technology/developers-tools/gemma-3/}{2025}]; $V=262{,}208$ & 2560 \\[3pt]
XGLM & --- & \href{https://huggingface.co/facebook/xglm-1.7B/blob/d23a5e8e2164af31a84a26756b9b17f925143050/config.json}{\nolinkurl{facebook/xglm-1.7B}}\newline [\href{https://arxiv.org/abs/2112.10668}{2021}]; $V=256{,}008$ & 2048 \\[3pt]
BLOOM & --- & \href{https://huggingface.co/bigscience/bloom-1b7/blob/cc72a88036c2fb937d65efeacc57a0c2ef5d6fe5/config.json}{\nolinkurl{bigscience/bloom-1b7}}\newline [\href{https://huggingface.co/bigscience/bloom-1b7}{2022}]; $V=250{,}880$ & 2048 \\[3pt]
Qwen & \href{https://huggingface.co/Qwen/Qwen-7B/blob/ef3c5c9c57b252f3149c1408daf4d649ec8b6c85/config.json}{\nolinkurl{Qwen/Qwen-7B}}\newline [\href{https://huggingface.co/Qwen/Qwen-7B}{2023}]; $V=151{,}936$ & \href{https://huggingface.co/Qwen/Qwen3.5-4B/blob/851bf6e806efd8d0a36b00ddf55e13ccb7b8cd0a/config.json}{\nolinkurl{Qwen/Qwen3.5-4B}}\newline [\href{https://qwen.ai/blog?id=qwen3.5}{2026}]; $V=248{,}320$ & 2560 \\[3pt]
Llama & \href{https://github.com/meta-llama/llama-models/blob/main/models/sku_list.py}{\nolinkurl{meta-llama/Llama-2-7b-hf}}\newline [\href{https://about.fb.com/news/2023/07/llama-2/}{2023}]; $V=32{,}000$ & \href{https://github.com/huggingface/transformers/blob/main/src/transformers/models/llama4/configuration_llama4.py}{\nolinkurl{meta-llama/Llama-4-Scout-17B-16E-Instruct}}\newline [\href{https://ai.meta.com/blog/llama-4-multimodal-intelligence/}{2025}]; $V=202{,}048$ & 5120 \\[3pt]
GPT & \href{https://huggingface.co/openai-community/gpt2/blob/607a30d783dfa663caf39e06633721c8d4cfcd7e/config.json}{\nolinkurl{openai-community/gpt2}}\newline [\href{https://huggingface.co/openai-community/gpt2}{2019}]; $V=50{,}257$ & \href{https://huggingface.co/openai/gpt-oss-20b/blob/6cee5e81ee83917806bbde320786a8fb61efebee/config.json}{\nolinkurl{openai/gpt-oss-20b}}\newline [\href{https://huggingface.co/openai/gpt-oss-20b}{2025}]; $V=201{,}088$ & 2880 \\[3pt]
Phi & \href{https://huggingface.co/microsoft/phi-2/blob/810d367871c1d460086d9f82db8696f2e0a0fcd0/config.json}{\nolinkurl{microsoft/phi-2}}\newline [\href{https://www.microsoft.com/en-us/research/?p=991293}{2023}]; $V=51{,}200$ & \href{https://huggingface.co/microsoft/Phi-4-mini-instruct/blob/cfbefacb99257ffa30c83adab238a50856ac3083/config.json}{\nolinkurl{microsoft/Phi-4-mini-instruct}}\newline [\href{https://huggingface.co/microsoft/Phi-4-mini-instruct}{2025}]; $V=200{,}064$ & 3072 \\[3pt]
GLM & \href{https://huggingface.co/THUDM/chatglm-6b/blob/bf0f5cfb575eebebf9b655c5861177acfee03f16/config.json}{\nolinkurl{THUDM/chatglm-6b}}\newline [\href{https://arxiv.org/html/2406.12793v1}{2023}]; $V=130{,}528$ & \href{https://huggingface.co/zai-org/GLM-4-9B-0414/blob/645b8482494e31b6b752272bf7f7f273ef0f3caf/config.json}{\nolinkurl{zai-org/GLM-4-9B-0414}}\newline [\href{https://github.com/zai-org/GLM-4}{2025}]; $V=151{,}552$ & 4096 \\[3pt]
Falcon & \href{https://huggingface.co/tiiuae/falcon-7b/blob/ec89142b67d748a1865ea4451372db8313ada0d8/config.json}{\nolinkurl{tiiuae/falcon-7b}}\newline [\href{https://huggingface.co/tiiuae/falcon-7b}{2023}]; $V=65{,}024$ & \href{https://huggingface.co/tiiuae/Falcon3-7B-Base/blob/bf3d7ed586cb22a921520e2d681a9d3d7642cde8/config.json}{\nolinkurl{tiiuae/Falcon3-7B-Base}}\newline [\href{https://huggingface.co/tiiuae/Falcon3-7B-Base}{2024}]; $V=131{,}072$ & 3072 \\[3pt]
Mistral & \href{https://huggingface.co/mistralai/Mistral-7B-v0.1/blob/27d67f1b5f57dc0953326b2601d68371d40ea8da/config.json}{\nolinkurl{mistralai/Mistral-7B-v0.1}}\newline [\href{https://mistral.ai/news/announcing-mistral-7b/}{2023}]; $V=32{,}000$ & \href{https://huggingface.co/mistralai/Ministral-3-3B-Instruct-2512-BF16/blob/b6d637bef2393152b3da2b2fde72eecdee30557e/config.json}{\nolinkurl{mistralai/Ministral-3-3B-Instruct-2512-BF16}}\newline [\href{https://mistral.ai/news/mistral-3/}{2025}]; $V=131{,}072$ & 3072 \\[3pt]
DeepSeek & \href{https://huggingface.co/deepseek-ai/deepseek-llm-7b-base/blob/7683fea62db869066ddaff6a41d032262c490d4f/config.json}{\nolinkurl{deepseek-ai/deepseek-llm-7b-base}}\newline [\href{https://github.com/deepseek-ai/DeepSeek-LLM/commits/main/}{2023}]; $V=102{,}400$ & \href{https://huggingface.co/deepseek-ai/DeepSeek-V3/blob/e815299b0bcbac849fa540c768ef21845365c9eb/config.json}{\nolinkurl{deepseek-ai/DeepSeek-V3}}\newline [\href{https://huggingface.co/deepseek-ai/DeepSeek-V3}{2024}]; $V=129{,}280$ & 7168 \\[3pt]
InternLM & \href{https://huggingface.co/internlm/internlm-7b/blob/96e127d08d851a88cac736a9b091dd953ae1b873/config.json}{\nolinkurl{internlm/internlm-7b}}\newline [\href{https://github.com/InternLM/InternLM}{2023}]; $V=103{,}168$ & \href{https://huggingface.co/internlm/internlm3-8b-instruct/blob/28c99415adaf61767bd1c619f4f99f308fdfd223/config.json}{\nolinkurl{internlm/internlm3-8b-instruct}}\newline [\href{https://github.com/InternLM/InternLM}{2025}]; $V=128{,}512$ & 4096 \\[3pt]
SmolLM & \href{https://huggingface.co/HuggingFaceTB/SmolLM-1.7B/blob/d7449ff7241c863f3e8accc475155f0f97afa011/config.json}{\nolinkurl{HuggingFaceTB/SmolLM-1.7B}}\newline [\href{https://huggingface.co/HuggingFaceTB/SmolLM-1.7B}{2024}]; $V=49{,}152$ & \href{https://huggingface.co/HuggingFaceTB/SmolLM3-3B/blob/a07cc9a04f16550a088caea529712d1d335b0ac1/config.json}{\nolinkurl{HuggingFaceTB/SmolLM3-3B}}\newline [\href{https://huggingface.co/HuggingFaceTB/SmolLM3-3B}{2025}]; $V=128{,}256$ & 2048 \\[3pt]
Hunyuan & --- & \href{https://huggingface.co/tencent/Hunyuan-A13B-Instruct/blob/290ddb9a56ed23c2c83a1c8081533e58925df952/config.json}{\nolinkurl{tencent/Hunyuan-A13B-Instruct}}\newline [\href{https://huggingface.co/tencent/Hunyuan-A13B-Instruct}{2025}]; $V=128{,}167$ & 4096 \\[3pt]
Baichuan & \href{https://huggingface.co/baichuan-inc/Baichuan-7B/blob/5d86e56a58fe4a5b3292cd9bb7468afef6f93eab/config.json}{\nolinkurl{baichuan-inc/Baichuan-7B}}\newline [\href{https://github.com/baichuan-inc/Baichuan-7B}{2023}]; $V=64{,}000$ & \href{https://huggingface.co/baichuan-inc/Baichuan2-7B-Base/blob/f9d4d8dd2f7a3dbede3bda3b0cf0224e9272bbe5/config.json}{\nolinkurl{baichuan-inc/Baichuan2-7B-Base}}\newline [\href{https://github.com/baichuan-inc/Baichuan2}{2023}]; $V=125{,}696$ & 4096 \\[3pt]
Granite & \href{https://huggingface.co/ibm-granite/granite-3.0-2b-base/blob/a8462c21f5e1f27be4536dc36d6cc789da23fbd6/config.json}{\nolinkurl{ibm-granite/granite-3.0-2b-base}}\newline [\href{https://huggingface.co/ibm-granite/granite-3.0-2b-base}{2024}]; $V=49{,}152$ & \href{https://huggingface.co/ibm-granite/granite-4.0-micro/blob/main/config.json}{\nolinkurl{ibm-granite/granite-4.0-micro}}\newline [\href{https://huggingface.co/ibm-granite/granite-4.0-micro}{2025}]; $V=100{,}352$ & 2560 \\[3pt]
StableLM & \href{https://huggingface.co/stabilityai/stablelm-base-alpha-3b/blob/819657db42861818efac86e8ca60f673fa912ef8/config.json}{\nolinkurl{stabilityai/stablelm-base-alpha-3b}}\newline [\href{https://github.com/Stability-AI/StableLM}{2023}]; $V=50{,}688$ & \href{https://huggingface.co/stabilityai/stablelm-2-1_6b/blob/f499ead74c53749bd93cebc6ce8bc0d7bdf1eaef/config.json}{\nolinkurl{stabilityai/stablelm-2-1_6b}}\newline [\href{https://stability.ai/news-updates/introducing-stable-lm-2}{2024}]; $V=100{,}352$ & 2048 \\[3pt]
OLMo & \href{https://huggingface.co/allenai/OLMo-7B-hf/blob/11fb3186a2e4f681edea621fa8b4345147a9db6a/config.json}{\nolinkurl{allenai/OLMo-7B-hf}}\newline [\href{https://huggingface.co/allenai/OLMo-7B-hf}{2024}]; $V=50{,}304$ & \href{https://huggingface.co/allenai/Olmo-3-7B-Think/blob/d97e442d7cc678210054dbcc9b440894d62c89a4/config.json}{\nolinkurl{allenai/Olmo-3-7B-Think}}\newline [\href{https://huggingface.co/allenai/Olmo-3-7B-Think}{2025}]; $V=100{,}278$ & 4096 \\[3pt]
RWKV & \href{https://huggingface.co/RWKV/rwkv-4-169m-pile/blob/46bdc280eb97b6141d5d51a935e0c4870ecaefcc/config.json}{\nolinkurl{RWKV/rwkv-4-169m-pile}}\newline [\href{https://huggingface.co/BlinkDL/rwkv-4-pile-169m}{2022}]; $V=50{,}277$ & \href{https://huggingface.co/RWKV/rwkv-5-world-1b5/blob/5e4ae736346d42e5c0fefaad5276271b04957aed/config.json}{\nolinkurl{RWKV/rwkv-5-world-1b5}}\newline [\href{https://huggingface.co/BlinkDL/rwkv-5-world/tree/main}{2023}]; $V=65{,}536$ & 2048 \\[3pt]
GPT-J & --- & \href{https://huggingface.co/EleutherAI/gpt-j-6b/blob/47e169305d2e8376be1d31e765533382721b2cc1/config.json}{\nolinkurl{EleutherAI/gpt-j-6b}}\newline [\href{https://huggingface.co/EleutherAI/gpt-j-6b}{2021}]; $V=50{,}400$ & 4096 \\[3pt]
OPT & --- & \href{https://huggingface.co/facebook/opt-6.7b/blob/a45aa65bbeb77c1558bc99bedc6779195462dab0/config.json}{\nolinkurl{facebook/opt-6.7b}}\newline [\href{https://huggingface.co/facebook/opt-6.7b}{2022}]; $V=50{,}272$ & 4096 \\[3pt]
StarCoder & --- & \href{https://huggingface.co/bigcode/starcoder2-3b/blob/733247c55e3f73af49ce8e9c7949bf14af205928/config.json}{\nolinkurl{bigcode/starcoder2-3b}}\newline [\href{https://huggingface.co/bigcode/starcoder2-3b}{2024}]; $V=49{,}152$ & 3072 \\[3pt]
OpenELM & --- & \href{https://huggingface.co/apple/OpenELM-3B/blob/f6eab253a25dbd1ac54340cc3ba44f9ec1ed5257/config.json}{\nolinkurl{apple/OpenELM-3B}}\newline [\href{https://huggingface.co/apple/OpenELM-3B}{2024}]; $V=32{,}000$ & 3072 \\[3pt]
\end{longtable}
\endgroup

%% file: generated/cross_model_likelihood.tex
\begin{tabular}{lrrrrr}
\toprule
Model & Head (M) & BF16 & W4 min--max & W4 AW-MSE & W8 AW-MSE \\
\midrule
Gemma 3 4B & 671 & 60.12 & 60.84 & 59.46 & 60.12 \\
Gemma 4 E4B & 671 & 74.97 & 75.48 & 75.55 & 74.95 \\
Qwen3.5 4B & 636 & 10.89 & 11.19 & 10.98 & 10.89 \\
Phi-4-mini & 615 & 11.65 & 40.58 & 30.94 & 11.75 \\
\bottomrule
\end{tabular}

%% file: sections/supp_confidence.tex
\section{Confidence and Likelihood Controls}
\label{app:confidence-controls}
\label{sec:temperature}

These unshifted-head controls distinguish preservation of source predictions from
incidental changes in confidence; they do not select a reparameterization
coefficient. Raw perplexity can reward a change in confidence rather than improved
source fidelity: in the original Gemma precision sweep, fitting separate source and
candidate temperatures on a held-out block half reverses apparent likelihood gains
(W4 AW-MSE excess NLL $-0.00364\!\to\!+0.04745$ nats/token; W3
$-0.04193\!\to\!+0.10236$), consistent with quantization softening an
overconfident source while adding distortion \citep{guo2017calibration}. This is
why we select representatives by source-to-candidate KL and report perplexity
separately. These controls concern the unshifted precision-sweep artifacts and therefore
do not establish temperature-calibrated likelihood behavior for the shifted heads.

%% file: sections/app_grouped.tex
\clearpage
\section{Higher-capacity parameterization of the equivalence class}
\label{app:grouped-extension}

\subsection{Parameterization and search}
The scalar parameterization can be generalized by assigning a separate
coefficient to each quantization group. Let $g(j)$ denote the group containing
hidden dimension $j$. We define
\begin{equation}
 W^{\mathrm{grp}}_{v,j}=W_{v,j}-t_{g(j)}\mu_j.
 \label{eq:grouped-reparam}
\end{equation}
Equivalently, $W^{\mathrm{grp}}=W-\mathbf1a^\top$ with
$a_j=t_{g(j)}\mu_j$, so the full-precision softmax distribution remains exactly
unchanged for a linear-softmax head. Aligning the coefficients with quantization
groups allows independently quantized regions of the head to select different
shift magnitudes while retaining the same packed layout and, for
shift-compatible logit paths, no additional inference operation.

The groups partition the hidden dimension; every vocabulary row receives the
same vector subtraction. With contiguous 128-channel groups, SmolLM3's
2,048-wide head has 16 coefficients. Setting every coefficient to the same $t$
recovers the scalar family. The extension enlarges the search space while the
matrix correction $\mathbf1a^\top$ remains rank one. Nonlinear paths use the
correction in Appendix~\ref{app:reparam-nonlinear}, restoring
$(a^\top h)\mathbf1$ before the logit transformation.

\paragraph{Optimization.}
We initialize all group coefficients at the validation-selected shared
$t^\star$ and perform coordinate-wise search over groups using validation KL.
At each update, we vary one coefficient while holding the others fixed and
retain a proposal only if it lowers full-vocabulary validation KL. The
coefficients are selected jointly through this objective, separately for each
base quantizer; they are frozen before test evaluation.

Groups are visited in descending order of $\sum_{j\in g}m_j\mu_j^2$,
using fitting moments $m_j=\E_{\rm fit}[h_j^2]$ only. We perform 32 coordinate
updates, each evaluating positive and negative proposals from the current
coefficient vector. The lower-KL proposal is accepted only if it improves on
the current vector. Step sizes are $0.5$, $0.125$ and $0.03125$ on successive
passes through the ordered groups, retaining the last step size for later
passes. The evaluated heads have 16--24 groups, so 32 updates use only the
first two step sizes and may stop partway through the second pass. Coefficients
are not clipped. GPTQ performs its own search, using its fitting Hessian and
validation objective.

We report two experiments with different search budgets and evaluation sets
(Table~\ref{tab:grouped-protocol}). Both initialization grids include zero
shift. The nine-model experiment extends the paper's 14-point scalar search
with 64 grouped proposals. The separate matched-budget SmolLM3 study compares grouped search
with scalar refinement using the same total candidate count. These finite
searches do not establish an optimal shift.

\begin{table}[ht]
\centering\small
\caption{Grouped-search protocols. Counts are validation evaluations per
quantizer. Each grouped search adds 32 updates with two proposals each.}
\label{tab:grouped-protocol}
\begin{tabular}{@{}p{0.27\textwidth}p{0.32\textwidth}p{0.32\textwidth}@{}}
\toprule
 & Nine-model experiment & Matched-budget SmolLM3 study \\
\midrule
Initialization & Main 14-point scalar grid & Shared 27-point scalar grid \\
Scalar budget & 14 & $27+64=91$ (scalar refinement) \\
Grouped budget & $14+64=78$ & $27+64=91$ \\
Evaluation & Main 16 test articles & 16 previously unused articles \\
Quantizers & RTN, AW-MSE, GPTQ & RTN, AW-MSE \\
\bottomrule
\end{tabular}
\end{table}

\clearpage
\subsection{Evaluation across nine models}
We evaluate W4 G128 heads from Phi-4-mini, Gemma 3/4, Qwen3.5, Qwen3,
Ministral, BLOOM, BLOOMZ and XGLM. The BF16 decoder remains fixed. English
WikiText supplies 128 fitting articles with eight states each, 16 validation
articles and 16 disjoint test articles, using 512-token prefixes. All choices
are frozen before test evaluation. This extension is evaluated on the main test
split and should be interpreted as exploratory rather than as an independent
replication.

Table~\ref{tab:grouped-models-kl} reports matched raw, scalar and grouped
results rerun together on H100. Grouped search lowers test KL on all nine
models under RTN and AW-MSE, with reductions of 3.0--20.3\% and 2.3--22.2\%,
respectively. For each quantizer, eight of nine paired intervals exclude zero.
GPTQ gains are smaller and mixed: five models improve, four regress slightly,
and only Phi's interval excludes zero. The additional validation budget
prevents attributing these gains solely to the larger parameterization.

\begin{table}[ht]
\centering\footnotesize
\setlength{\tabcolsep}{3pt}
\caption{\textbf{Grouped search lowers RTN/AW-MSE KL on all nine models; GPTQ results are mixed.}
W4 G128 KL uses each model's source; scalar/grouped budgets are 14/78 validation
evaluations. Positive reductions favor grouped search;
$\Delta=\mathrm{KL}_{\rm grp}-\mathrm{KL}_{\rm scalar}$.
Pointwise paired article-bootstrap 95\% intervals use 10,000 resamples,
conditional on frozen selections, on previously examined test articles.
Gemma~4 ($\dagger$) restores the common offset before soft-capping.}
\label{tab:grouped-models-kl}
\input{generated/grouped_models_kl.tex}
\end{table}

All metrics use reconstructed heads, full-vocabulary FP32 projection and
log-softmax on frozen BF16 states, with TF32 disabled. RTN uses signed
$-7,\ldots,7$ codes and FP32 scales; AW-MSE and GPTQ use signed $-8,\ldots,7$
codes, the 12-point clipping grid, and BF16 scales and reconstruction.
GPTQ uses 1\% damping and no activation ordering
(Appendix~\ref{app:base-quantizers}). Gemma 4 uses the original pretrained
checkpoint and restores $(a^\top h)\mathbf1$ before its soft-cap with threshold
30. These results do not measure packed-serving latency.

Table~\ref{tab:grouped-models-ppl} gives perplexity on the same test states.
KL reduction does not imply lower perplexity in every configuration; for
example, Gemma 3 AW-MSE improves KL while slightly increasing perplexity.
The source references and numerical path belong to this comparison and should
not be pooled with the packed-deployment measurements.

\begin{table}[H]
\centering\small
\caption{\textbf{Lower KL does not always yield lower perplexity.}
Nine-model scalar $\to$ grouped perplexity uses the same states as
Table~\ref{tab:grouped-models-kl}. Source PPL is each model's BF16 reference.}
\label{tab:grouped-models-ppl}
\input{generated/grouped_models_ppl.tex}
\end{table}

\subsection{SmolLM3 comparison with matched candidate counts}
The separate matched-budget SmolLM3 study initializes both searches from 27 scalar candidates.
Grouped search spends 64 additional evaluations on the coordinate proposals
above; the scalar baseline spends 64 on finer one-dimensional refinement.
Both therefore use 91 validation evaluations, although their search regions
and optimization procedures differ.

On 16 previously unused English WikiText articles, the frozen grouped shift
reduces SmolLM3 W4 G128 KL from 0.8923 to 0.8008 under RTN and from 0.3404 to
0.2848 under AW-MSE, relative to the refined scalar search. These articles come
from the WikiText validation corpus split and exclude the main fitting,
validation, and test articles. Paired article-bootstrap 95\% intervals for
$\mathrm{KL}_{\rm grp}-\mathrm{KL}_{\rm 1D}$ are $[-0.1232,-0.0630]$ and
$[-0.0702,-0.0414]$, respectively (10,000 resamples). This is an exploratory
single-model matched-budget comparison. The measurements use
reconstructed heads on matched frozen decoder states; packed-serving latency
was not measured for the grouped extension.

%% file: generated/grouped_models_kl.tex
\begin{tabular}{@{}llrrrrrl@{}}
\toprule
Model & Quantizer & $t^\star$ & Raw & Scalar & Grouped & Red. (\%) & $\Delta$ KL: 95\% CI \\
\midrule
Phi-4-mini & RTN & 4 & 1.230021 & 0.350579 & 0.300249 & $+14.36$ & $[-0.059382,-0.042322]$ \\
 & AW-MSE & 4 & 0.939398 & 0.255508 & 0.222474 & $+12.93$ & $[-0.040437,-0.026940]$ \\
 & GPTQ & 4 & 0.160414 & 0.060511 & 0.058634 & $+3.10$ & $[-0.003556,-0.000413]$ \\
Gemma 3 & RTN & 0.5 & 0.049208 & 0.046005 & 0.038335 & $+16.67$ & $[-0.009860,-0.005429]$ \\
 & AW-MSE & -1 & 0.040232 & 0.037523 & 0.033956 & $+9.51$ & $[-0.005063,-0.002085]$ \\
 & GPTQ & 0 & 0.033609 & 0.033609 & 0.032296 & $+3.91$ & $[-0.003080,+0.000345]$ \\
Gemma 4$^\dagger$ & RTN & 1 & 0.011617 & 0.010587 & 0.010273 & $+2.97$ & $[-0.000662,+0.000044]$ \\
 & AW-MSE & 0.5 & 0.008402 & 0.008248 & 0.007313 & $+11.33$ & $[-0.001141,-0.000728]$ \\
 & GPTQ & 1 & 0.008423 & 0.008132 & 0.008329 & $-2.42$ & $[-0.000021,+0.000399]$ \\
Qwen3.5 & RTN & -0.5 & 0.020831 & 0.018588 & 0.016096 & $+13.41$ & $[-0.002920,-0.002062]$ \\
 & AW-MSE & -0.5 & 0.013097 & 0.013075 & 0.011662 & $+10.81$ & $[-0.001813,-0.001038]$ \\
 & GPTQ & -0.5 & 0.010844 & 0.010470 & 0.010629 & $-1.52$ & $[-0.000146,+0.000456]$ \\
Qwen3 & RTN & -0.5 & 0.028519 & 0.027787 & 0.023311 & $+16.11$ & $[-0.005166,-0.003727]$ \\
 & AW-MSE & 0 & 0.013296 & 0.013296 & 0.011808 & $+11.19$ & $[-0.001841,-0.001141]$ \\
 & GPTQ & 0 & 0.007929 & 0.007929 & 0.008129 & $-2.52$ & $[-0.000079,+0.000477]$ \\
Ministral & RTN & 0 & 0.007443 & 0.007443 & 0.007061 & $+5.13$ & $[-0.000534,-0.000251]$ \\
 & AW-MSE & 0 & 0.005576 & 0.005576 & 0.005450 & $+2.26$ & $[-0.000284,+0.000053]$ \\
 & GPTQ & 0 & 0.006135 & 0.006135 & 0.006097 & $+0.63$ & $[-0.000178,+0.000093]$ \\
BLOOM & RTN & 6 & 1.116508 & 0.531289 & 0.423429 & $+20.30$ & $[-0.136209,-0.085733]$ \\
 & AW-MSE & 4 & 0.594330 & 0.136469 & 0.121634 & $+10.87$ & $[-0.018924,-0.011113]$ \\
 & GPTQ & 4 & 0.036659 & 0.028343 & 0.028132 & $+0.74$ & $[-0.001123,+0.000791]$ \\
BLOOMZ & RTN & 4 & 1.119557 & 0.483042 & 0.385792 & $+20.13$ & $[-0.109115,-0.085646]$ \\
 & AW-MSE & 4 & 0.662490 & 0.157715 & 0.122734 & $+22.18$ & $[-0.039459,-0.029717]$ \\
 & GPTQ & 4 & 0.041895 & 0.032715 & 0.032838 & $-0.38$ & $[-0.000105,+0.000337]$ \\
XGLM & RTN & 1 & 2.128650 & 0.143396 & 0.127086 & $+11.37$ & $[-0.022235,-0.011954]$ \\
 & AW-MSE & 1 & 0.586145 & 0.094916 & 0.085099 & $+10.34$ & $[-0.013016,-0.006262]$ \\
 & GPTQ & 1 & 0.008968 & 0.006694 & 0.006691 & $+0.04$ & $[-0.000136,+0.000141]$ \\
\bottomrule
\end{tabular}

%% file: generated/grouped_models_ppl.tex
\begin{tabular}{@{}lrrrr@{}}
\toprule
 & & \multicolumn{3}{c}{Scalar $\to$ grouped PPL} \\
\cmidrule(lr){3-5}
Model & Source PPL & RTN & AW-MSE & GPTQ \\
\midrule
Phi-4-mini & 9.719 & 13.461$\to$13.042 & 12.486$\to$12.106 & 10.286$\to$10.287 \\
Gemma 3 & 37.977 & 38.482$\to$38.419 & 38.086$\to$38.176 & 38.852$\to$38.613 \\
Gemma 4$^\dagger$ & 67.223 & 68.156$\to$68.011 & 68.192$\to$68.129 & 67.646$\to$67.961 \\
Qwen3.5 & 8.977 & 9.131$\to$9.143 & 9.106$\to$9.060 & 9.065$\to$9.054 \\
Qwen3 & 12.577 & 13.020$\to$12.839 & 12.747$\to$12.617 & 12.636$\to$12.639 \\
Ministral & 8.598 & 8.629$\to$8.633 & 8.665$\to$8.662 & 8.660$\to$8.632 \\
BLOOM & 18.469 & 29.826$\to$27.845 & 21.030$\to$20.774 & 18.868$\to$18.864 \\
BLOOMZ & 22.067 & 31.237$\to$29.965 & 25.271$\to$25.145 & 22.745$\to$22.670 \\
XGLM & 13.809 & 15.877$\to$15.712 & 15.483$\to$15.120 & 13.886$\to$13.862 \\
\bottomrule
\end{tabular}